\documentclass[sigconf]{aamas}

\usepackage{balance} 
\usepackage{amsmath}
\usepackage{algorithm}
 \usepackage{algorithmicx} \usepackage{algpseudocode}
\usepackage{graphicx}
\usepackage{microtype}
\usepackage{wrapfig}
\usepackage{hyperref}       
\usepackage{url}            
\usepackage{booktabs}       
\usepackage{amsfonts}       
\usepackage{nicefrac}       
\usepackage{microtype}      
\usepackage{xcolor}         
\usepackage{enumitem}  
\usepackage{xspace}    
\usepackage{enumitem}
\usepackage{subcaption}
\usepackage[font=small,skip=1pt]{caption}
\usepackage{float}      
\usepackage{wrapfig}    
\usepackage{xcolor}
\usepackage{soul}
\definecolor{myblue}{rgb}{0.1,0.2,0.7}

\setcopyright{ifaamas}
\acmConference[AAMAS '26]{Proc.\@ of the 25th International Conference
on Autonomous Agents and Multiagent Systems (AAMAS 2026)}{May 25 -- 29, 2026}
{Paphos, Cyprus}{C.~Amato, L.~Dennis, V.~Mascardi, J.~Thangarajah (eds.)}
\copyrightyear{2026}
\acmYear{2026}
\acmDOI{}
\acmPrice{}
\acmISBN{}

\acmSubmissionID{525}

\title{PREFINE: Preference-Based Implicit Reward and Cost Fine-Tuning for Safety Alignment}

\author{Richa Verma}
\affiliation{
  \institution{TCS Research, \\Department of CSE, IIT Madras} \country{India}}
\email{richa.verma4@tcs.com}

\author{Bavish Kulur}
\affiliation{
  \institution{Department of Computing Science, \\University of Alberta} \country{Canada}}
\email{bavish@ualberta.ca}

\author{Sanjay Chawla}
\affiliation{
  \institution{Qatar Computing Research Institute, \\Hamad Bin Khalifa University} \country{Qatar}}
\email{schawla@hbku.edu.qa}

\author{Balaraman Ravindran}
\affiliation{
  \institution{Department of Data Science \& AI, \\ Wadhwani School of
Data Science \& AI, IIT Madras} \country{India}}
\email{ravi@dsai.iitm.ac.in}

\begin{abstract}
We address the problem of making a pre-trained reinforcement learning (RL) policy safety-aware by incorporating cost constraints without retraining it from scratch. While costs could be numerically encoded, we assume a more general setting is when costs are provided as preferences. Given a reward-optimized policy and a small dataset of preferred (low-cost) and dispreferred (high-cost) trajectories, our goal is to fine-tune the policy to generate low-cost behaviors while retaining high rewards.  Unlike standard RLHF in language models, where preferences are defined over responses to the same prompt, our setting involves trajectory-level preferences in continuous control environments. We introduce \textbf{\textsc{PREFINE}}: \textbf{Pre}ference-based Implicit Reward and Cost \textbf{Fine}-Tuning for Safety Alignment which is a preference-based fine-tuning method that adapts \textbf{Direct Preference Optimization (DPO)}, which is now widely used for LLM fine-tuning, to the sequential decision making setting. \textsc{PREFINE} constructs policy-sampled counterfactual trajectories to establish meaningful preference contrasts and jointly optimizes for reward retention and safety alignment. Empirically, \textsc{PREFINE} reduces constraint violations and catastrophic failures by over 60\% while maintaining original reward behavior. \textsc{PREFINE} produces policies that achieve low-cost, high-reward performance with significantly improved data and computational efficiency compared to full offline RL or imitation learning, bridging preference alignment and safe policy adaptation in continuous domains.
\end{abstract}


\newcommand{\BibTeX}{\rm B\kern-.05em{\sc i\kern-.025em b}\kern-.08em\TeX}

\begin{document}


\pagestyle{fancy}
\fancyhead{}

\maketitle 
\section{Introduction}
\label{sec:intro}
As the focus in machine learning and AI shifts from prediction to prescription, safety of AI applications is now of paramount importance. In RL \cite{sutton1998}, safety constraints are modeled by extending the MDP with cost constraints and using Lagrangian methods ~\cite{achiam2017constrained, xu2022constraints, stooke2020responsive, liu2022constrained} to solve the resulting constrained optimization problem. However, this approach is not computationally feasible as it involves solving a nested optimization problem. Likewise, standard imitation learning (IL) ~\cite{argall2009survey, ross2011reduction, schaal1996learning} focuses solely on replicating expert demonstrations and disregards non‑preferred, unsafe trajectories, leaving no mechanism to enforce safety constraints. Instead, there is a growing demand for solutions that can leverage offline data and fine-tune a policy to meet safety constraints. Inspired by the work in RLHF in LLMs ~\cite{christiano2017deep}, we propose using an extension of Direct Preference Optimization (DPO) to incorporate safety constraints while ensuring that the \textit{safety alignment} is fully offline.
\begin{figure*}[ht]
\centering
\includegraphics[width=0.8\textwidth,height=5.6cm]{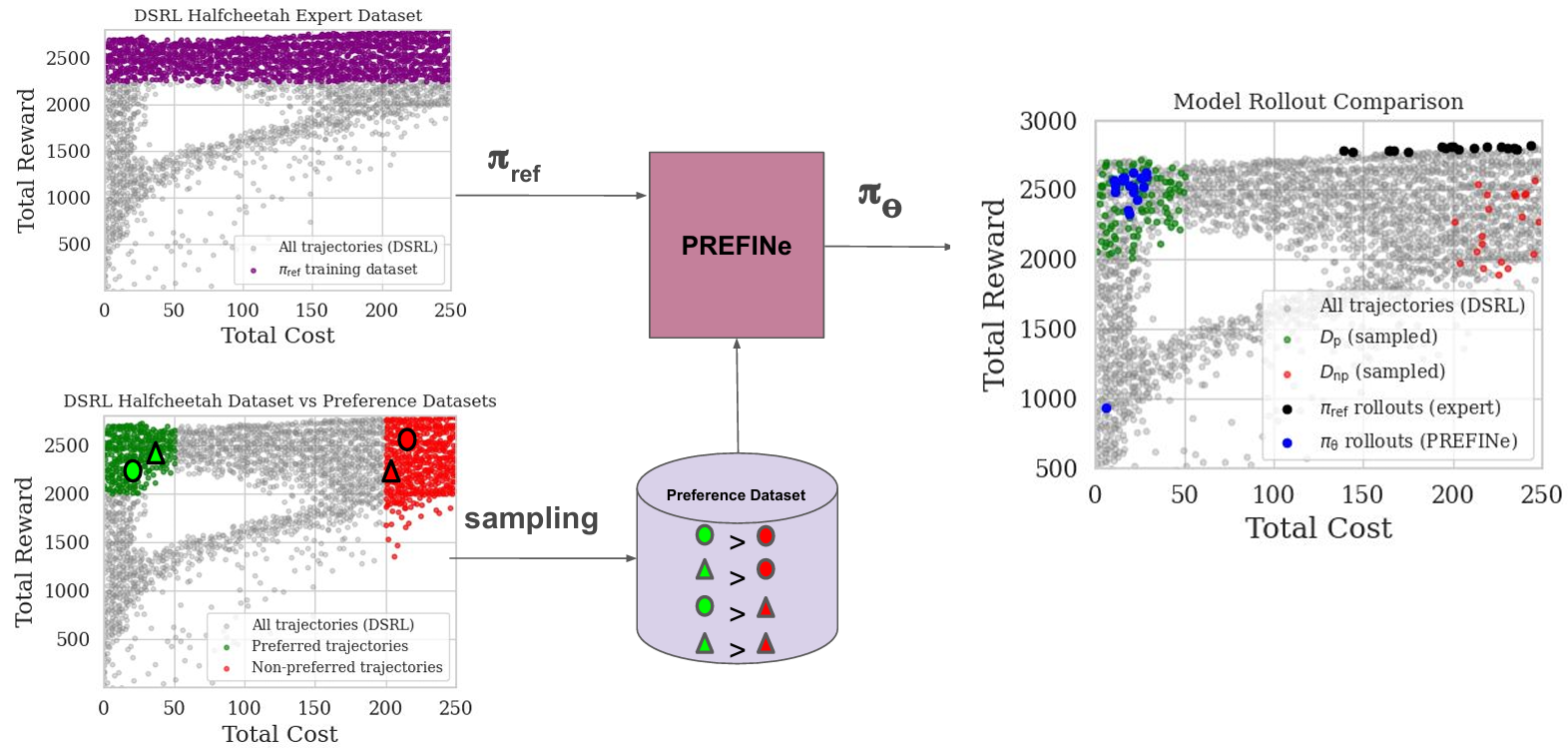}
\small
\caption{Overview of the \textsc{PREFINE} pipeline. (\textbf{Top-left}) The DSRL HalfCheetah offline dataset (grey) contains trajectories with a wide range of costs and rewards; we pre-train a reference policy $\pi_{\text{ref}}$ on the high-reward, low-cost subset (purple). (\textbf{Bottom-left}) We sample a small \emph{preferred} set $\mathcal{D}_p$ (green) of safe trajectories and a \emph{non-preferred} set $\mathcal{D}_{np}$ (red) of unsafe trajectories to form pairwise comparisons. (\textbf{Center}) \textsc{PREFINE} ingests $\pi_{\text{ref}}$ and these preference pairs, then fine-tunes in a single-stage DPO–SFT loop to produce a new policy $\pi_{\theta}$. (\textbf{Right}) Rollouts of $\pi_{\theta}$ (blue) shift into the low-cost, high-reward region, retaining the performance of original $\pi_{\text{ref}}$ rollouts (black) and avoiding unsafe behaviors (red) without any online interaction.}
\label{fig:teaser}
\end{figure*}
We introduce \textbf{\textsc{PREFINE}}: \textbf{Pre}ference-based Implicit Reward \textbf{Fine}-Tuning for Safety Alignment, a fully offline framework that transforms safety alignment into a preference-based learning problem. Rather than requiring hand‑crafted cost signals or reward information, \textsc{PREFINE} relies on pairwise comparisons between safe (preferred) and unsafe (non‑preferred) trajectories to implicitly encode safety without any online interaction. We combine Direct Preference Optimization (DPO) ~\cite{rafailov2023dpo} with supervised fine‑tuning (SFT) ~\cite{ouyang2022training} to incorporate cost constraints into any differentiable pre‑trained policy (RL or IL) without explicit cost estimation. DPO is widely used to fine-tune Large Language Models (LLMs) using human preferences. We adapt it to the sequential decision making setting in two steps: (i) we create a set of counterfactual trajectories using an existing pre-trained policy which share common states with the given trajectories in the preference dataset, and (ii) we use both reward and preference optimization to fine-tune the policy. Note that in a real world setting, such as autonomous driving, unsafe (or high cost) data is hard to collect. Therefore, \textsc{PREFINE} uses a mix of abundant safe (preferred) data and scarce unsafe (non-preferred) data to perform safety alignment. We show an overview of our approach in Figure ~\ref{fig:teaser}. The expert policy $\pi_{ref}$ learns from the high-reward region of the underlying DSRL ~\cite{liu2024dsrl} dataset (see Figure ~\ref{fig:teaser}(Top-left). \textsc{PREFINE} leverages mixed-quality datasets spanning the reward-cost Pareto frontier. Safe (i.e. preferred) demonstrations are marked in green while the unsafe (i.e. non-preferred) demonstrations have been highlighted in red (see Figure ~\ref{fig:teaser}(Bottom-left)). In Figure ~\ref{fig:teaser}(Right), we see that expert policy ($\pi_{ref}$) rollouts show that it achieves high rewards but exhibits unsafe cost variations (black). \textsc{PREFINE} learns the policy $\pi_{safe}$ and it is evident from the rollouts that $\pi_{safe}$ avoids high-cost regions while retaining expert-level performance (blue).

Our hybrid DPO+SFT objective mirrors the Regularized Preference Optimization ~\cite{NEURIPS2024_fa69e968} principle i.e., combining a preference optimization term with an imitation (SFT) regularizer to mitigate over-optimization, formalized for RLHF settings. We adapt this structure to sequential decision-making with offline safety preferences, using SFT to anchor policy updates under partial coverage while optimizing preference-consistent comparisons. Our approach has several advantages and novel features including (i) We show that we can create policies which result in low-cost trajectories while maintaining high rewards and (ii) \textsc{PREFINE} is substantially more 
cost (time) efficient and (iii) we work in a setting that is neither imitation learning nor complete offline RL.

The main contributions of the work are as follows:
\begin{itemize}
\item We formulate safety alignment as a preference-based learning problem, using trajectory preferences, enabling fully offline policy updates for retrofitting cost constraints into existing RL policies.
\item We introduce \textsc{PREFINE}, a novel approach to combine DPO's preference ranking with SFT's stability that enables direct policy learning while eliminating the need for challenging min-max optimization. We adapt DPO (for the first time) from LLMs to safety alignment in RL. 
\item We show that \textsc{PREFINE} reduces constraint violations and catastrophic failures by 60\% - 92\% while retaining expert‑level task performance, and converges with a wall-clock time an order of magnitude smaller compared to cost‑based and distribution‑matching baselines.
\end{itemize}
For evaluation, we use the well-established DSRL benchmark \cite{liu2024dsrl}, a safe offline RL evaluation benchmark. We present the results of extensive experiments that we conducted to compare our method with several state-of-the-art baselines in terms of normalized reward and normalized cost across 12 continuous control tasks in two popular domains. Note that preserving policy performance in terms of reward is prioritized over minimizing cost, which here serves as a proxy for safety. 
\textsc{PREFINE} is designed for scenarios where costs are better defined as preferences and multiple decision trajectories achieve the same goal but differ in safety, such as autonomous driving or surgical interventions. In driving, for instance, reaching a destination is the objective, yet overspeeding makes the trajectory unsafe. Similarly, a surgery may succeed overall, even if certain intermediate actions are risky. In such settings, humans can more naturally express relative safety preferences (e.g., “which driving trajectory is safer?”) than assign precise numerical safety scores (e.g., “this driving trajectory rates 7.3/10”). \textbf{We show that \textsc{PREFINE} successfully learns policies in the setting where costs are not quantitatively defined but expressed as preferences and it retains the high rewards of the pre-trained reference policy while becoming safety-aware, outperforming the state-of-the-art baselines.} We demonstrate that \textsc{PREFINE} provides a fast, scalable recipe for retrofitting pre‑trained policies with safety (cost constraints) using \textit{safety alignment}, paving the way for a broader real-world deployment of RL systems. \textsc{PREFINE} code and datasets are available here \footnote{https://github.com/zxXhVi/\textsc{PREFINE}}.
\section{Preliminaries}
In principle, safe RL problems 
are often modeled as
using \textit{Constrained Markov Decision Processes} (CMDPs)~\cite{altman1999constrained}, defined by a tuple $\mathcal{M} = (\mathcal{S}, \mathcal{A}, T, r, c, \mu_0, \gamma)$, where $\mathcal{S}$ is the state space, $\mathcal{A}$ is the action space, $T: \mathcal{S} \times \mathcal{A} \rightarrow \Delta(\mathcal{S})$ describes the transition dynamics, $r: \mathcal{S} \times \mathcal{A} \rightarrow \mathbb{R}$ is the reward function, and $c: \mathcal{S} \times \mathcal{A} \rightarrow [0, C_{\text{max}}]$ is the cost function, where $C_{\text{max}}$ denotes the maximum allowable cost. The distribution $\mu_0: \mathcal{S} \rightarrow [0,1]$ represents the initial state distribution, and $\gamma \in [0, 1)$ is the discount factor. Let $\pi(a \mid s)$ denote the probability of taking action $a$ in state $s$ under policy $\pi$. The reward-based state-value function is defined as $V_r^{\pi}(s) = \mathbb{E}_{\pi} \left[ \sum_{t=0}^{\infty} \gamma^t r(s_t, a_t) \mid s_0 = s \right]$, and the cost-based value function as $V_c^{\pi}(s) = \mathbb{E}_{\pi} \left[ \sum_{t=0}^{\infty} \gamma^t c(s_t, a_t) \mid s_0 = s \right]$.

While CMDPs assume access to explicit cost signals, in many real-world safety-critical domains such costs are unavailable or hard to specify. Instead, we have \textbf{pairwise preferences} between trajectories, reflecting which behaviors are safer and desirable. Preference-based learning provides a way to incorporate such feedback.
\subsection{Reinforcement Learning from Human Feedback (RLHF)}
In reinforcement learning from human feedback (RLHF), the goal is to align a policy $\pi_\theta$ with human preferences using a dataset of pairwise comparisons  
$\mathcal{D} = \{(x, y_w, y_l)\}$,  
where $x$ is a prompt or context, $y_w$ is a preferred response, and $y_l$ is a less preferred one.  
A common probabilistic formulation for such comparisons is the \textbf{Bradley--Terry model}~\cite{bradley1952rank}, which expresses the probability of preferring $y_w$ over $y_l$ as a softmax over their latent rewards:
\small
\begin{equation}
    \Pr[y_w \succ y_l \mid x] =
    \frac{\exp\!\big(r_E(x,y_w)\big)}{
          \exp\!\big(r_E(x,y_w)\big) + \exp\!\big(r_E(x,y_l)\big)},
\end{equation}
\normalsize
where $r_E(x,y)$ denotes an implicit reward function. This provides a probabilistic link between preference data and latent rewards. \\
\textbf{Direct Preference Optimization (DPO)}~\cite{rafailov2023dpo} builds on this by directly optimizing the policy $\pi_\theta$ with respect to preference data, without learning $r_E$ explicitly.  
Given $(x, y_w, y_l)$, DPO minimizes:
\small
\begin{equation}
    \mathcal{L}_{\text{DPO}}(\pi_\theta;\pi_{\text{ref}}) =
    -\,\log\sigma\!\Big(
    \beta\big[
      \log\tfrac{\pi_\theta(y_w|x)}{\pi_{\text{ref}}(y_w|x)} -
      \log\tfrac{\pi_\theta(y_l|x)}{\pi_{\text{ref}}(y_l|x)}
    \big]\Big),
\end{equation}
\normalsize
where $\pi_{\text{ref}}$ is a reference policy and $\beta$ controls preference sharpness.  
This objective implicitly defines a reward model  
$\hat{r}(x,y) = \beta\log \frac{\pi_\theta(y|x)}{\pi_{\text{ref}}(y|x)}$,  
encouraging $\pi_\theta$ to assign higher likelihood to preferred responses while remaining close to $\pi_{\text{ref}}$. \\
\textbf{Supervised Fine-Tuning (SFT)} complements this by maximizing the likelihood of desirable responses:
\small
\begin{equation}
    \mathcal{L}_{\text{SFT}}(\pi_\theta)
    = -\,\mathbb{E}_{(x,y_w)\sim\mathcal{D}}\!
      \big[\log \pi_\theta(y_w|x)\big],
\end{equation}
\normalsize
which stabilizes optimization by anchoring the policy on preferred data, although it does not explicitly penalize undesirable outputs.

In this work, we adapt these methods, which are originally developed for aligning large language models with human preferences, to the offline sequential decision-making setting. Bradley-Terry model provides the statistical foundation for modeling preferences, DPO introduces a principled contrastive loss against a reference policy, and SFT anchors learning to safe demonstrations. Together, these components form the basis for our preference-guided fine-tuning framework for sequential decision making.
\begin{figure}[t]
    \centering
    \begin{subfigure}[t]{0.9\linewidth}
        \centering
        \includegraphics[width=0.8\linewidth, height=60mm]{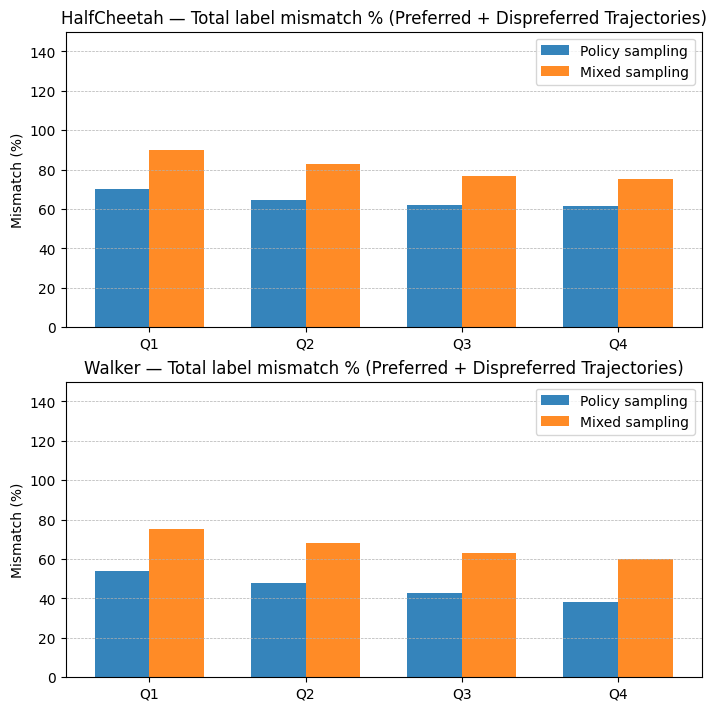}
    \end{subfigure}
    \vspace{-1mm}
    \caption{\textbf{Evolution of total label mismatch percentage across training quartiles.}  
Each bar denotes the fraction of pairwise comparisons exhibiting contradictory binary cost labels between dataset and counterfactual trajectories (preferred + dispreferred combined). Results are shown for HalfCheetah and Walker tasks under two counterfactual sampling strategies: policy-sampled (\textsc{PREFINE}) and mixed (dataset+policy).  
\textsc{PREFINE} consistently yields fewer mismatches and a clearer downward trend over training, indicating reduced label noise and more coherent preference supervision as fine-tuning progresses. In contrast, mixed sampling produces persistently higher mismatch rates, reflecting noisier counterfactual comparisons and weaker safety alignment.}

    \label{fig:cost_gap_pref_dispref}
\end{figure}
\section{\textsc{PREFINE}}
In this section, we describe \textsc{PREFINE}’s formulation as a preference-based fine-tuning procedure for post-hoc safety alignment in an offline sequential decision-making setting. We present a hybrid loss combining Direct Preference Optimization (DPO)~\cite{rafailov2023dpo} with supervised fine-tuning (SFT)~\cite{ouyang2022training}, and the training algorithm.
\subsection{Problem Formulation} 
We consider the task of \emph{offline post-hoc safety alignment}, where the goal is not to learn a safe policy from scratch, but to \emph{retrofit an existing policy} $\pi_{\text{ref}}$ with safety (or cost constraints). The reference policy $\pi_{\text{ref}}$ is assumed to be a differentiable policy trained beforehand (via RL or IL) to achieve high reward.  The environment is modeled as a constrained Markov Decision Process (CMDP), defined by reward $r(s,a)$, cost $c(s,a)$, and a cost threshold $\tau$. A trajectory is considered \emph{safe enough} if its cumulative cost lies below $\tau$.\\
\textbf{Problem Definition:} Given a reference policy $\pi_{ref}$, a preference data set $D_{p}$ of high reward and low cost trajectories from $\pi_{ref}$ and a data set $D_{np}$ of
non-preferred high reward and high cost trajectories. Let $\phi$ be a trajectory and let
$P_{\theta}(\phi)$ be the probability of selecting a trajectory $\phi$ when using policy
$\pi_{\theta}$. Let $\tau$ be the cost threshold. The objective is to finetune $\pi_{ref}$ using cost preferences in an offline manner such that
the resulting $\pi_{\theta}$ satisfies the following:
\begin{align}
\underset{\phi \sim P_{{\theta}}}{\mathbf{E}}[R(\phi)] \equiv \underset{\phi \sim P_{ref}}{\mathbf{E}}[R(\phi)] \\
\underset{\phi \sim P_{{\theta}}}{\mathbf{E}}[c(\phi)] \leq \tau
\end{align}
This distinguishes our setting from imitation learning (which learns directly from demonstrations) and offline RL (which learns from scratch with batch data and focuses on reward maximization). Instead, we formulate \textsc{\textsc{PREFINE}} as a \emph{preference-based policy refinement problem}, aligning $\pi_{\text{ref}}$ with safety preferences while retaining its original reward capabilities. This means that \textsc{PREFINE} starts from a strong policy $\pi_{\text{ref}}$ and only wants to reduce the costs without sacrificing the reward. This hybrid positioning is coherent: \textsc{PREFINE} uses costs defined as preferences (safe vs unsafe actions) rather than explicit costs, and fine-tunes a base policy rather than re-training from scratch. \textbf{Note that the "cost" is just a proxy for the notion of safety. PREEFINe is designed to work in settings where costs are better expressed as preferences e.g., safe vs. rash driving by an autonomous vehicle, as explained in Section\ref{sec:intro}}.
We achieve this through a \emph{hybrid preference optimization objective}, combining Direct Preference Optimization (DPO) to align with pairwise cost preferences, and a supervised fine-tuning (SFT) anchor for reward retention. The trajectory dataset $\mathcal{D}$ is partitioned into preferred (safe) trajectories $D_p$ and non-preferred (unsafe) trajectories $D_{np}$ based on the cost threshold and the reward. \textsc{PREFINE} then aligns $\pi_{\text{ref}}$ with $\pi_\theta$ using pairwise cost preferences derived from $(D_p, D_{np})$, while ensuring reward retention through supervised fine-tuning and without any environment interaction. This \emph{post-hoc offline safety alignment} ensures $\pi_\theta$ reduces cost constraint violations without sacrificing the rewards.\\
\textbf{Dynamic sampling strategy:} We propose a novel way to apply DPO, a preference optimization loss function from i.i.d. language modeling setting, to the dynamic, sequential decision-making context of Reinforcement Learning. DPO is used for LLM fine-tuning with a preference dataset consisting of triples of the format <\textit{prompt, preferred response, non-preferred response}>. To map DPO to the sequential decision-making setting, for a given state $s$ (analogous to a user prompt) and the associated action $a^+$ (analogous to a preferred response) in the static offline preference dataset, \textsc{PREFINE} generates a counterfactual action $a^-$ for the state-action pair $(s,a^+) \sim \mathcal{D}_p$ to construct per-state action triples $<s, a^+, a^->$. The resulting data triples are further utilized by a DPO-inspired loss function. We explain this in detail in Section \ref{sec:action-pairing}.\\
\textbf{Data and notation:}
We assume offline trajectory logs partitioned into:
(i) $\mathcal{D}_{p}$: safe trajectories with low cost and high reward (preferred), and
(ii) $\mathcal{D}_{np}$: unsafe trajectories with high cost (non-preferred).
We extract state-action pairs from these for training.
\subsection{Practical Implementation for Trajectory Datasets}
\label{sec:action-pairing}
Computing the DPO loss theoretically requires a preference pair per state, that is, $<s, a^+, a^->$ ~\cite{rafailov2023dpo}. For the counterfactual action in a preference pair, a natural choice is to sample directly from the datasets whenever possible. Concretely, when we wish to draw a preferred pair \((s,a^{+})\) from \(D_{p}\), we can locate the nearest state \(s'\) in the opposite dataset (\(D_{np}\)) and use the paired action of that state as counterfactual action (\(a^{-}\)). However, in practice, exact matches of high-dimensional states across offline trajectory datasets (e.g., DSRL) are often unavailable. Moreover, if dataset nearest-neighbor counterfactuals are often unrealistic/off-distribution, they produce weak or misleading contrasts. This is further exacerbated by our chosen setting of keeping the size of $\mathcal{D}_{np}$ as small as possible to maintain similarity with real-world scenarios. As a workaround, we sample counterfactual actions from the current policy $\pi_{\theta}$ simply because policy-sampled alternatives are closer to the decision boundary and therefore more informative. 
\textbf{Justification for using policy-sampled counterfactuals:} 
Given a trajectory $\mathcal{T} = \{(s_1, a_1), (s_2, a_2),...,(s_T, a_T)\}$, we want to generate a contrasting trajectory $\mathcal{T'} = \{(s_1, a'_1), (s_2, a'_2),...,(s_T, a'_T)\}$, where the set of actions $\{a'_i\} \sim \pi_{\theta}$, such that $c(\mathcal{T'}) > c(\mathcal{T})$ if $\mathcal{T} \sim \mathcal{D}_{p}$ and $c(\mathcal{T'}) < c(\mathcal{T})$ if $\mathcal{T} \sim \mathcal{D}_{np}$. Sampling the set of counterfactual actions from the current policy $\pi_{\theta}$ ensure that those actions reflect actual deployment behavior and creates a progressively harder set of contrasts as $\pi_{\theta}$ improves, thereby improving the safety signal. Note that the cost of each individual action $a'_i$ sampled from $\pi_{\theta}$ may not always be higher (when dataset action $a_i \sim \mathcal{D}_p$) or lower ($a_i \sim \mathcal{D}_{np}$) and this may introduce noise in the training process. To avoid the label noise introduced in this manner, using a supervised fine-tuning (SFT) anchor to the reference policy to preserve reward capability helps and it acts as an implicit regularizer. Our ablation (Fig.~\ref{fig:ab1_app}) shows that removing the SFT anchor leads to catastrophic drops in performance, corroborating its necessity. This sampling strategy creates an adaptive self-critique signal where the policy is continually compared against alternatives it would actually propose. Essentially, it sharpens learning as \(\pi_\theta\) improves and the sampled actions become more informative. \emph{Note that it is desired that the only the expected cost of a counterfactual trajectory $\mathcal{T}$ is higher, when $\mathcal{T} \sim \mathcal{D}_p$, and lower when $\mathcal{T} \sim \mathcal{D}_{np}$.} \textbf{For the sake of completion, we provide a comparison of the percentage of noisy (false) labels encountered for policy-sampling and dataset-based nearest neighbour with policy fallback when dataset actions are not available (mixed) sampling strategies in Figure ~\ref{fig:cost_gap_pref_dispref}.}\\
\textbf{Interpretation of training-phase trends:}
Figure\ref{fig:cost_gap_pref_dispref} reports the evolution of the \emph{total label-mismatch percentage} (preferred + dispreferred binary cost mismatches) computed per training quartile over the first 200K fine-tuning steps. A "binary label flip" indicates that the cost (safe/unsafe) indicator differs between the dataset trajectory and the counterfactual trajectory sampled from the current policy for the same state. The plots show a consistent pattern on both HalfCheetahVelocity and Walker2dVelocity: when counterfactuals are sampled from the current policy (i.e., \textsc{PREFINE}), mismatch rates begin high and decline during training and we see HalfCheetahVelocity $\approx70\%\rightarrow\approx62\%$, Walker $\approx54\%\rightarrow\approx38\%$, indicating that pairwise comparisons become more coherent as the policy learns safer actions. By contrast, the mixed (dataset+policy fallback) sampling strategy exhibits substantially larger and more persistent mismatch rates (HalfCheetahVelocity $\approx90\%\rightarrow\approx75\%$, Walker2dVelocity $\approx75\%\rightarrow\approx60\%$). We attribute this systematic excess of contradictions under mixed sampling to label noise introduced by dataset-sampled counterfactuals: dataset actions can be stale or distributionally mismatched relative to the evolving policy, which produces contradictory preference labels and corrupts the supervision signal. In summary, policy-sampled counterfactuals produce a cleaner, more stable preference signal. 

Together, these complementary trends demonstrate that the policy sampling-based counterfactual action generation mechanism successfully differentiates between safe and unsafe state regions thereby improving safety and reducing the average number of cost constraint violations.
These conclusions assume that the offline dataset affords reasonable local coverage of the policy's state distribution. The preferred and nonpreferred trajectory sets exhibit roughly \(70\%\) Jaccard similarity across our tasks and UMAP visualizations (Figure \ref{fig:umap}) show substantial overlap in visited states.
\subsection{Hybrid DPO+SFT Objective with Policy-sampled Actions}
\label{sec:loss}
We optimize a hybrid objective that combines a DPO-style preference loss with an SFT anchor. Let $\pi_{\text{ref}}$ be the reference policy. Our implemented loss is as follows:
\small
\begin{align}
\Delta(s,a^+,a^-)
&= \log \tfrac{\pi_\theta(a^+|s)}{\pi_{\text{ref}}(a^+|s)}
 - \log \tfrac{\pi_\theta(a^-|s)}{\pi_{\text{ref}}(a^-|s)} .
\end{align}
\vspace{-3pt}
\begin{align}
L_p(\pi_\theta;\pi_{\text{ref}}) 
&= -\,\mathbb{E}_{(s,a^+)\sim\mathcal{D}_p,\,a^-\sim\pi_\theta}\!\Big[
\log\sigma\!\big(\beta\,\Delta(s,a^+,a^-)\big) \nonumber\\[-2pt]
&\hspace{1.5cm} +\, \lambda\,\log\pi_\theta(a^+|s)\Big], \label{eq:lp}\\[-1pt]
L_{np}(\pi_\theta;\pi_{\text{ref}}) 
&= -\,\mathbb{E}_{(s,a^-)\sim\mathcal{D}_{np},\,a^+\sim\pi_\theta}\!
\Big[\log\sigma\!\big(\beta\,\Delta(s,a^+,a^-)\big)\Big], \label{eq:lnp}\\[-1pt]
\mathcal{L}_{\text{\textsc{PREFINE}}}(\pi_\theta;\pi_{\text{ref}}) 
&= L_p(\pi_\theta;\pi_{\text{ref}}) + L_{np}(\pi_\theta;\pi_{\text{ref}}).
\label{eq:prefine-impl}
\end{align}
The first expectation in Eq.\ref{eq:lp} pushes up the relative logit of known-safe $a^+$ against policy-sampled counterfactual action $a^-$, while the SFT term $\lambda \log \pi_\theta(a^+|s)$ anchors reward retention; the second expectation in Eq.\ref{eq:lnp} pushes down known-unsafe $a^-$ against policy-sampled counterfactual action $a^+$. Here, $\lambda (>0)$ balances safety alignment against reward retention and $\beta (>0)$ controls the strength of KL-divergence in the original DPO loss function~\cite{rafailov2023dpo}. Empirically, $\beta$ and $\lambda$ are selected via validation set to achieve the desired safety–performance trade-off. 

This operationalizes a regularized preference optimization pattern analogous to RPO (DPO loss + SFT loss combination) ~\cite{NEURIPS2024_fa69e968}. The combined objective,\(\mathcal{L}_{\text{\textsc{PREFINE}}} = \mathcal{L}_{\text{DPO}} + \lambda \mathcal{L}_{\text{SFT}},
\)
is functionally identical to RPO’s \(\mathcal{L}_{\text{RPO}} = \mathcal{L}_{\text{DPO}} + \eta \beta \cdot \mathcal{L}_{\text{SFT}}\), with \(\lambda\) playing the role of \(\eta \beta\). Consequently, \textsc{PREFINE} aligns structurally with and is motivated by RPO. While RPO's theory is developed for language models with bandit feedback, we adopt its structural insight for fine-tuning with offline cost preferences in sequential control and empirically validate its benefits here.

In RPO, the SFT loss is a regularizer that prevents the policy from exploiting gaps in the static data coverage (i.e., over-optimization). In Prefine, it serves a dual, even more critical purpose: (i) It acts as a powerful anchor that stabilizes the entire learning process. By constantly pulling the policy $\pi_{\theta}$ towards high-reward behaviors of policy $\pi_{ref}$. (ii) It ensures that the actions $a^+$ or $a^-$ sampled from $\pi_{\theta}$ remain reasonable. Without the SFT anchor, $\pi_{\theta}$ could drift into a state where the generated actions might be too noisy, completely corrupting the DPO learning signal. We show the importance of including the SFT term in \textsc{PREFINE}'s loss function in Fig. \ref{fig:ab1_app}.
\subsection{Efficient Single-Stage Training Procedure}The fine-tuning procedure outlined in Algorithm~\ref{alg:prefine_alg} requires only a single backward pass per batch i.e., no nested loops or separate models for estimating reward and cost. We construct per-state counterfactual actions via policy sampling (as discussed in Section~\ref{sec:action-pairing}) and optimize Eq.~\eqref{eq:prefine-impl}.
This single‑stage update circumvents the need for nested optimization loops, yielding up to $10\times$ less wall-clock time than cost-based methods (as shown in Section \ref{sec:experiments}), making \textsc{PREFINE} practical for large-scale offline safety-critical tasks. For the exact implementation details, see Appendix.
\small
\begin{algorithm}[t]
\caption{\textsc{PREFINE} — policy-sampled counterfactual trajectories (algpseudocode)}
\label{alg:prefine_alg}
\begin{algorithmic}[1]
\Require Preferred trajectory dataset $D_p$, non-preferred trajectory dataset $D_{np}$, reference policy $\pi_{\mathrm{ref}}$ (frozen), learnable policy $\pi_\theta$, batch size $B$, trajectory length $T$, hyperparameters $\beta,\lambda,\eta$
\State Pretrain $\pi_{\mathrm{ref}}$ via BC on offline dataset and keep it fixed.
\For{each training iteration}
  \State Sample minibatch of $B$ preferred trajectories $\{\mathcal{T}^p_i\}_{i=1}^B$ from $D_p$ \& $B$ non-preferred trajectories $\{\mathcal{T}^{np}_j\}_{j=1}^B$ from $D_{np}$.
  \State{$\mathcal{T}^p_i=\{(s_{i,t},a^+_{i,t})\}$, $\mathcal{T}^{np}_j=\{(s_{j,t},a^-_{j,t})\}$}
  \State Initialize empty sets $\mathcal{B}_p \leftarrow \varnothing$ and $\mathcal{B}_{np} \leftarrow \varnothing$.
  \For{each preferred trajectory $\mathcal{T}^p_i$}
    \For{$t\leftarrow 1$ \textbf{to} $T$}
        \State Sample counterfactual action $a^{-}_{i,t} \sim \pi_\theta(\cdot\mid s_{i,t})$.
        \State Add triple $(s_{i,t},\,a^+_{i,t},\,a^{-}_{i,t})$ to $\mathcal{B}_p$.
    \EndFor
  \EndFor
  \For{each non-preferred trajectory $\mathcal{T}^{np}_j$}
    \For{$t\leftarrow 1$ \textbf{to} $T$}
        \State Sample counterfactual action $a^{+}_{j,t} \sim \pi_\theta(\cdot\mid s_{j,t})$.
        \State Add triple $(s_{j,t},\,a^{+}_{j,t},\,a^-_{j,t})$ to $\mathcal{B}_{np}$.
    \EndFor
  \EndFor
  \ForAll{$(s,a^+,a^-)\in\mathcal{B}_p \cup \mathcal{B}_{np}$} \Comment{Compute for every triple}
    \State $\Delta(s,a^+,a^-)\gets \log\frac{\pi_\theta(a^+\mid s)}{\pi_{\mathrm{ref}}(a^+\mid s)}
      - \log\frac{\pi_\theta(a^-\mid s)}{\pi_{\mathrm{ref}}(a^-\mid s)}$.
  \EndFor
  \State Compute \textsc{PREFINE} loss as per Eq. \ref{eq:prefine-impl}:
  \State $\theta \gets \theta - \eta \nabla_\theta L_{\text{\textsc{PREFINE}}}$  \Comment{Gradient step (Adam)}
\EndFor
\end{algorithmic}
\end{algorithm}
\section{Related Work}
\label{sec:related_work}
Several recent lines of work address safe offline RL and imitation learning via preferences, distribution matching, or explicit constraint modeling.  \textbf{SafeDPO}~\cite{kim2025safedpo} adapts Direct Preference Optimization with cost-based regularization, but it is fundamentally different from \textsc{\textsc{PREFINE}}: SafeDPO depends on explicit cost signals and is not designed for the highly imbalanced, offline regimes we target. By contrast, the DICE family (e.g., \textbf{SafeDICE}~\cite{jang2023safedice}, \cite{kostrikov2020valuedice,kim2022demodice}) performs stationary distribution matching with learned cost critics and nested or adversarial optimizations—making them natural and meaningful baselines for offline imitation-style objectives, but computationally heavier than our single-stage fine-tuning. For these reasons we evaluate against SafeDICE as a strong offline baseline.

\textbf{TraC}~\cite{gong2025offline} uses the same benchmark but addresses a different problem: it actively rebalances preference datasets via relabeling and querying to obtain near-uniform coverage of preferred/dispreferred behaviors. \textsc{\textsc{PREFINE}} explicitly assumes a more challenging setting of fixed, highly \emph{unbalanced} offline dataset with very few unsafe trajectories and performs preference-based fine-tuning without any active data curation; hence a direct comparison to TraC is not warranted because the methods operate under different assumptions and evaluation goals.

Other recent approaches—e.g., latent safety-constrained policy learning~\cite{koirala2024latent} and constraint-conditioned actor-critic methods~\cite{guo2025constraint} seek to encode safety in latent spaces or via constraint-conditioned optimization. \textsc{\textsc{PREFINE}} differs from these by requiring only a small number of unsafe demonstrations, and using a single-stage fine-tuning objective without explicit cost estimation that both preserves task performance and enforces safety alignment efficiently. Together, these distinctions motivate our baseline choices.
\small
\begin{figure*}[!htbp]
        \centering
        \includegraphics[width=\linewidth]{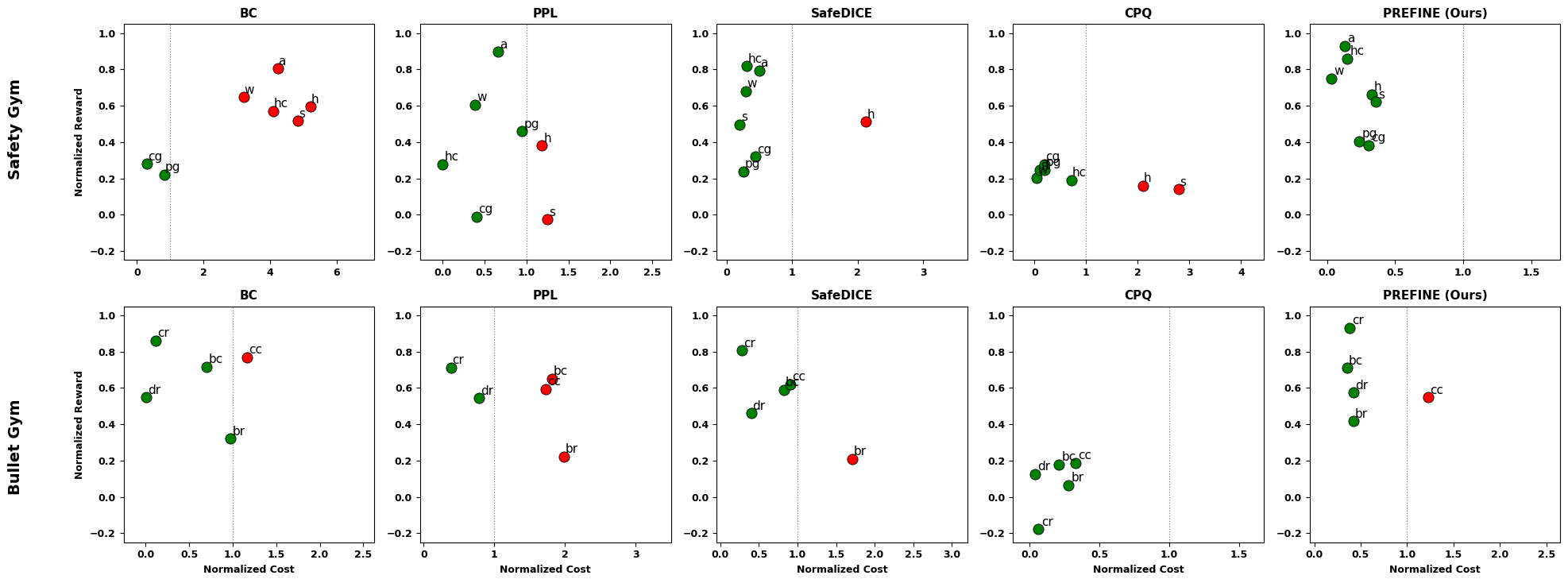}
    \small
    \caption{Comparison of \textsc{PREFINE} against baselines in Safety Gym (top) and Bullet Gym (bottom). Each dot denotes a task; green indicates satisfaction of the safety constraint (normalized cost $\leq$ 1), while red indicates a violation. The vertical dotted line corresponds to the normalized cost threshold of 1. \textsc{PREFINE} consistently concentrates points in the top-left region (high reward, low cost), whereas baselines either violate constraints (BC, PPL, SafeDICE) or trade-off reward for cost (CPQ). Each dot has an abbreviated task label mapped to full task names in Table \ref{tab:results}.}

    \label{fig:main_res}
\end{figure*}
\begin{table*}[t]
\centering
\begin{subtable}[t]{\linewidth}
\centering
\caption*{Safety Gym}
\resizebox{\linewidth}{!}{%
\begin{tabular}{l r r r r r r r r r r r r r}
\toprule
Task & \multicolumn{2}{c}{Reference policy ($\pi_{\text{ref}}$)} & \multicolumn{2}{c}{BC} & \multicolumn{2}{c}{PPL} & \multicolumn{2}{c}{SafeDICE} & \multicolumn{2}{c}{CPQ} & \multicolumn{2}{c}{\textsc{PREFINE} (Ours)} \\
\cmidrule(lr){2-3}\cmidrule(lr){4-5}\cmidrule(lr){6-7}\cmidrule(lr){8-9}\cmidrule(lr){10-11}\cmidrule(lr){12-13}
& Reward {\scriptsize$\uparrow$} & Cost {\scriptsize$\downarrow$} &
Reward {\scriptsize$\uparrow$} & Cost {\scriptsize$\downarrow$} &
Reward {\scriptsize$\uparrow$} & Cost {\scriptsize$\downarrow$} &
Reward {\scriptsize$\uparrow$} & Cost {\scriptsize$\downarrow$} &
Reward {\scriptsize$\uparrow$} & Cost {\scriptsize$\downarrow$} &
Reward {\scriptsize$\uparrow$} & Cost {\scriptsize$\downarrow$} \\
\midrule
AntVelocity \textbf{(a)} & 0.93&8.04 & 0.81 $\pm$ 0.05 & 4.23 $\pm$ 0.06 & \textbf{0.90 $\pm$ 0.04} & \textbf{0.66 $\pm$ 0.02} & \textbf{0.79 $\pm$ 0.04} & \textbf{0.50 $\pm$ 0.02} & \textbf{0.24 $\pm$ 0.05} & \textbf{0.10 $\pm$ 0.09} & \textcolor{blue}{\textbf{0.93 $\pm$ 0.04}} & \textcolor{blue}{\textbf{0.13 $\pm$ 0.03}} \\
CarGoal \textbf{(cg)} & 0.44 & 3.69 & \textbf{0.28 $\pm$ 0.06} & \textbf{0.31 $\pm$ 0.06} & \textbf{-0.01 $\pm$ 0.05} & \textbf{0.40 $\pm$ 0.02} & \textbf{0.32 $\pm$ 0.05} & \textbf{0.44 $\pm$ 0.07} & \textbf{0.28 $\pm$ 0.01} & \textbf{0.20 $\pm$ 0.04} & \textcolor{blue}{\textbf{0.38 $\pm$ 0.01}} & \textcolor{blue}{\textbf{0.30 $\pm$ 0.01}} \\
HalfCheetahVelocity \textbf{(hc)} & 0.99 & 8.01 & 0.57 $\pm$ 0.01 & 4.10 $\pm$ 0.04 & \textbf{0.28 $\pm$ 0.01} & \textbf{0.00 $\pm$ 0.06} & \textbf{0.70 $\pm$ 0.05} & \textbf{0.31 $\pm$ 0.04} & \textbf{0.19 $\pm$ 0.03} & \textbf{0.71 $\pm$ 0.03} & \textcolor{blue}{\textbf{0.86 $\pm$ 0.05}} & \textcolor{blue}{\textbf{0.15 $\pm$ 0.33}} \\
HopperVelocity \textbf{(h)} & 0.93 & 8.24 & 0.60 $\pm$ 0.00 & 5.22 $\pm$ 0.04 & 0.38 $\pm$ 0.02 & 1.19 $\pm$ 0.03 & 0.51 $\pm$ 0.08 & 2.12 $\pm$ 0.03 & 0.16 $\pm$ 0.05 & 2.10 $\pm$ 0.08 & \textcolor{blue}{\textbf{0.66 $\pm$ 0.03}} & \textcolor{blue}{\textbf{0.33 $\pm$ 0.05}} \\
PointGoal \textbf{(pg)} & 0.64 & 2.82 & \textbf{0.22 $\pm$ 0.05} & \textbf{0.82 $\pm$ 0.05} & \textcolor{blue}{\textbf{0.46 $\pm$ 0.01}} & \textcolor{blue}{\textbf{0.95 $\pm$ 0.05}} & \textbf{0.24 $\pm$ 0.09} & \textbf{0.26 $\pm$ 0.00} & \textbf{0.25 $\pm$ 0.08} & \textbf{0.20 $\pm$ 0.05} & \textbf{0.40 $\pm$ 0.03} & \textbf{0.24 $\pm$ 0.11} \\
SwimmerVelocity \textbf{(s)} & 0.94 & 6.34 & 0.52 $\pm$ 0.07 & 4.83 $\pm$ 0.01 & -0.03 $\pm$ 0.05 & 1.25 $\pm$ 0.07 & \textbf{0.50 $\pm$ 0.04} & \textbf{0.20 $\pm$ 0.07} & 0.14 $\pm$ 0.05 & 2.79 $\pm$ 0.07 & \textcolor{blue}{\textbf{0.62 $\pm$ 0.09}} & \textcolor{blue}{\textbf{0.36 $\pm$ 0.03}} \\
Walker2dVelocity \textbf{(w)} & 0.93 & 9.81 & 0.65 $\pm$ 0.06 & 3.21 $\pm$ 0.04 & \textbf{0.61 $\pm$ 0.05} & \textbf{0.39 $\pm$ 0.01} & \textbf{0.68 $\pm$ 0.03} & \textbf{0.29 $\pm$ 0.04} & \textbf{0.20 $\pm$ 0.08} & \textbf{0.04 $\pm$ 0.08} & \textcolor{blue}{\textbf{0.75 $\pm$ 0.01}} & \textcolor{blue}{\textbf{0.03 $\pm$ 0.04}} \\
\bottomrule
\end{tabular}
}
\end{subtable}
\begin{subtable}[t]{\linewidth}
\centering
\caption*{Bullet Gym}
\resizebox{\linewidth}{!}{%
\begin{tabular}{l r r r r r r r r r r r r r}
\toprule
Task & \multicolumn{2}{c}{Reference policy ($\pi_{\text{ref}}$)} & \multicolumn{2}{c}{BC} & \multicolumn{2}{c}{PPL} & \multicolumn{2}{c}{SafeDICE} & \multicolumn{2}{c}{CPQ} & \multicolumn{2}{c}{\textsc{PREFINE} (Ours)} \\
\cmidrule(lr){2-3}\cmidrule(lr){4-5}\cmidrule(lr){6-7}\cmidrule(lr){8-9}\cmidrule(lr){10-11}\cmidrule(lr){12-13}
& Reward {\scriptsize$\uparrow$} & Cost {\scriptsize$\downarrow$} &
Reward {\scriptsize$\uparrow$} & Cost {\scriptsize$\downarrow$} &
Reward {\scriptsize$\uparrow$} & Cost {\scriptsize$\downarrow$} &
Reward {\scriptsize$\uparrow$} & Cost {\scriptsize$\downarrow$} &
Reward {\scriptsize$\uparrow$} & Cost {\scriptsize$\downarrow$} &
Reward {\scriptsize$\uparrow$} & Cost {\scriptsize$\downarrow$} \\
\midrule
BallCircle \textbf{(bc)} &0.89&3.73  & \textcolor{blue}{\textbf{0.72 $\pm$ 0.07}} & \textcolor{blue}{\textbf{0.70 $\pm$ 0.01}} & 0.65 $\pm$ 0.04 & 1.82 $\pm$ 0.05 & \textbf{0.59 $\pm$ 0.09} & \textbf{0.82 $\pm$ 0.07} & \textbf{0.18 $\pm$ 0.04} & \textbf{0.21 $\pm$ 0.04} & \textbf{0.71 $\pm$ 0.04} & \textbf{0.35 $\pm$ 0.10} \\
BallRun \textbf{(br)} & 0.95& 3.88& \textbf{0.32 $\pm$ 0.04} & \textbf{0.97 $\pm$ 0.04} & 0.22 $\pm$ 0.06 & 1.99 $\pm$ 0.05 & 0.21 $\pm$ 0.04 & 1.71 $\pm$ 0.02 & \textbf{0.07 $\pm$ 0.05} & \textbf{0.28 $\pm$ 0.04} & \textcolor{blue}{\textbf{0.41 $\pm$ 0.08}} & \textcolor{blue}{\textbf{0.41 $\pm$ 0.07}} \\
CarRun \textbf{(cr)} & 0.99&1.89 & \textbf{0.86 $\pm$ 0.05} & \textbf{0.11 $\pm$ 0.07} & \textbf{0.71 $\pm$ 0.04} & \textbf{0.39 $\pm$ 0.08} & \textbf{0.81 $\pm$ 0.05} & \textbf{0.28 $\pm$ 0.07} & \textbf{-0.18 $\pm$ 0.01} & \textbf{0.06 $\pm$ 0.06} & \textcolor{blue}{\textbf{0.93 $\pm$ 0.04}} & \textcolor{blue}{\textbf{0.38 $\pm$ 0.01}} \\
CarCircle \textbf{(cc)} & 0.97& 4.58 & 0.77 $\pm$ 0.04 & 1.17 $\pm$ 0.02 & 0.59 $\pm$ 0.05 & 1.73 $\pm$ 0.12 & \textcolor{blue}{\textbf{0.62 $\pm$ 0.03}} & \textcolor{blue}{\textbf{0.91 $\pm$ 0.03}} & \textbf{0.19 $\pm$ 0.06} & \textbf{0.33 $\pm$ 0.06} & 0.54 $\pm$ 0.04 & 1.23 $\pm$ 0.04 \\
DroneRun \textbf{(dr)} & 0.84 & 6.77& \textbf{0.55 $\pm$ 0.07} & \textbf{0.01 $\pm$ 0.07} & \textbf{0.54 $\pm$ 0.02} & \textbf{0.79 $\pm$ 0.05} & \textbf{0.46 $\pm$ 0.05} & \textbf{0.40 $\pm$ 0.02} & \textbf{0.13 $\pm$ 0.06} & \textbf{0.04 $\pm$ 0.02} & \textcolor{blue}{\textbf{0.57 $\pm$ 0.08}} & \textcolor{blue}{\textbf{0.42 $\pm$ 0.06}} \\
\bottomrule
\end{tabular}
}
\end{subtable}
\caption{\textbf{Comprehensive per-task results.} Normalized reward (higher is better) and normalized cost (lower is better). Blue indicates the \emph{highest-reward safe agent} per task; \textbf{bold} marks agents with cost less than the cost threshold $<1$.}
\label{tab:results}
\end{table*}
\section{Experiments}
\label{sec:experiments}
In this section, we present an extensive empirical study conducted for \textsc{PREFINE} across various safety-critical control tasks. We specifically address the following three key questions:

\textbf{Q1 Effectiveness:} Can \textsc{PREFINE} meaningfully reduce cost violations compared to other baselines, while preserving the task performance? 

\textbf{Q2 Efficiency:} How quickly does \textsc{PREFINE} converge in terms of wall-clock time relative to the baseline methods?

\textbf{Q3 Robustness \& Sensitivity:} How does \textsc{PREFINE} respond to preferred datasets ($|\mathcal{D}_{p}|$) collected at various cost thresholds $\tau$? How does \textsc{PREFINE}’s performance vary with the size of preferred dataset ($|\mathcal{D}_{p}|$), non-preferred dataset ($|\mathcal{D}_{np}|$), and the choice of important hyperparameters.
\subsection{Experimental Setup}
For our experiments, we use the well-established DSRL benchmark \cite{liu2024dsrl} which provides trajectory datasets with wide-ranging rewards and costs for offline safe RL. DSRL datasets have been collected by training multiple policies with SOTA safe RL algorithms with varying cost thresholds. Each dataset consists of a mix of trajectories ranging from safe to highly unsafe generated by different policies, which makes it multimodal in nature. The offline datasets have been collected for 38 tasks across widely recognized environments. We choose 12 tasks out of those due to compute limitations: 7 from SafetyGym ~\cite{ji2023safety} and 5 from BulletSafetyGym \cite{gronauer2022bullet}. To evaluate performance, we follow the constraint variation evaluation protocol proposed in DSRL \cite{liu2024dsrl}, which is designed to test the adaptability of different algorithms. Specifically, we run each algorithm on every dataset under three separate cost thresholds and across five random seeds, allowing for a fair comparison. The evaluation is based on normalized reward and normalized cost \cite{liu2024dsrl}, where achieving a normalized cost value below 1 signifies that the policy is safe. In the case of implementing \textsc{PREFINE}, we begin by pretraining an expert policy through behavior cloning (BC) using high-reward (irrespective of the cost) trajectories of the offline dataset (as shown in Figure \ref{fig:teaser}), which serves as the reference policy $\pi_{ref}$ to be used for safety alignment. Next, \textsc{PREFINE} is applied to learn a new policy from the constructed sets of desirable and undesirable trajectories. \textsc{PREFINE} is agnostic to the choice of the underlying training method as long as $\pi_{ref}$ is differentiable in nature. We use variational autoencoder (VAE) architecture ~\cite{kingma2014auto} with \textsc{PREFINE}, mainly to capture the multimodality of the DSRL datasets.\\
\textbf{Dataset construction:} All trajectory subsets (preferred $D_{p}$ and non-preferred $D_{np}$) are constructed by a deterministic, pre-specified protocol applied to the same offline task-specific datasets provided by DSRL benchmark suite \cite{liu2024dsrl}. For a fixed cost threshold $\tau$ we first split the corpus into $\text{SAFE} = \{\tau_i \mid \text{cumulative\_cost}(\tau_i) < \tau\}$ and $\text{UNSAFE} =  \{\tau_i \mid \text{cumulative\_cost}(\tau_i) \geq \tau\}$. To form $D_{p}$, we select $N_{p}$ trajectories from SAFE by stratified sampling across the top reward quantiles (ties resolved by dataset index); to form $D_{np}$ we sample $N_{np}$ trajectories from the bottom 100 trajectories from UNSAFE (sorted by cost) so that $D_{np}$ spans the reward range. The values $N_{p}$ and $N_{np}$ are fixed prior to model training (we use $N_{p}=100$ and $N_{np}=20$) and the sampling code is published with the released code. We ran experiments with multiple values of $N_{p}$ and $N_{np}$, and finally chose the smallest ones that consistently gave good results in multiple tasks. \emph{Note that we present results averaged across three values of $\tau$ and 5 different sampled datasets $D_p$ and $D_{np}$ for each value of $\tau$ to avoid any bias in dataset construction.} Also note that not all actions within a "safe" trajectory are individually safe; our state-level objective works with expected cost of a trajectory, with SFT regularization mitigating noise from occasional mislabeled steps.\\
\textbf{Baselines:} We compare against the following baselines: (1) BC: Behaviour Cloning trained on trajectories from preferred dataset $\mathcal{D}_{p}$ that satisfy safety constraints. (2) PPL (Bradley-Terry preference learning)~\cite{bradley1952rank}: Direct preference learning without SFT anchoring initialized by the reference policy $\pi_{ref}$ for a fair comparison. (3) SafeDICE~\cite{jang2023safedice}: Offline safe imitation learning via distribution matching and cost critics. SafeDICE performs distribution correction explicitly and uses a mixture of unlabelled and labelled trajectory datasets to learn safe behaviour. We modify our datasets in the same way to check SafeDICE performance fairly. (4) CPQ~\cite{xu2022constraints}: a Q-learning based approach that treats unsafe actions as out-of-distribution actions and penalizes them. The hyperparameters for all baselines follow their original publications; code and seeds are fixed across methods to ensure fair comparison. We perform hyperparameter tuning for \textsc{PREFINE} to choose the values of $\beta$ and $\lambda$ and use the same values for each group of tasks i.e., SafetyGym and BulletSafetyGym, respectively (see Appendix for more). We intentionally skip comparing \textsc{PREFINE} against other relevant methods such as DWBC~\cite{wu2022dwbc} and DExperts~\cite{mitchell2022dexperts} because SafeDICE already outperforms those ~\cite{jang2023safedice}. All the experiments are run on an NVIDIA A100 GPU. The results are averaged over 5 different sampled datasets across 3 values of the cost threshold ($\tau$), five random seeds and 100 rollouts at each evaluation step.
\subsection{Effectiveness (Q1)}
Figure \ref{fig:main_res} contrasts normalized reward versus normalized cost across tasks. Points to the left of the vertical line (normalized cost = 1) satisfy the safety constraint. \textsc{PREFINE} concentrates in the upper-left region on both suites, indicating higher reward and lower cost relative to baselines. In Safety Gym, \textsc{PREFINE} is safe on all tasks while achieving the highest rewards; in Bullet Gym it is safe on most tasks with competitive rewards, exhibiting only a single violation. In contrast, BC and PPL often violate constraints, SafeDICE shows occasional unsafe outliers with reduced reward, and CPQ attains safety (low cost) by substantially under-optimizing reward. These results support our claim that preference-guided fine-tuning from policy samples yields a policy that is safe without being over-conservative.

The main results are presented in Table \ref{tab:results}, which reports the performance of all approaches in terms of normalized reward and normalized cost across 12 tasks averaged over 5 seeds and 3 cost thresholds. \textsc{PREFINE} consistently attains the best trade-off between reward and cost. In Safety Gym, \textsc{PREFINE} achieves the highest average normalized reward (except on PointGoal) while keeping costs well below the safety threshold, clearly outperforming all baselines. In Bullet Gym, \textsc{PREFINE} matches or exceeds the reward levels of the baselines while reducing violations substantially (except on BallRun and CarCircle). When averaged across environments, \textsc{PREFINE} delivers the strongest overall performance. In contrast, other baselines either incur frequent violations (BC, PPL, SafeDICE in Safety Gym) or achieve only marginal rewards while remaining conservative (CPQ).

Regarding safety constraint satisfaction, we present the proportion of tasks solved safely by each approach in Figure \ref{fig:safe_percent}  (see Appendix). \textsc{PREFINE} consistently achieves the highest fraction of safe tasks across environments, reaching 100\% in Safety Gym and 80\% in Bullet Gym. In contrast, while baselines such as SafeDICE and CPQ demonstrate moderate levels of safety (83\% and 71\% in Safety Gym; 60\% and 100\% in Bullet Gym, respectively), they either fail to generalize across both environments or sacrifice reward excessively to remain safe. Other approaches, including BC and PPL, show much lower safe task fractions and frequently violate cost thresholds. These results highlight that \textsc{PREFINE} is able to combine broad safety coverage with competitive rewards, unlike baselines that trade one for the other.

These results demonstrate \textsc{PREFINE}’s ability to retain expert-level behavior while dramatically improving safety.  It manages to achieve that without learning a reward model explicitly (like PPL), performing distribution correction and learning a cost model explicitly (like SafeDICE) or performing complex nested optimization to learn a seperate cost model (like CPQ), and without any online environment interactions. This reduction in cost is achieved with minimal loss in task reward. This balance between performance and safety arises from our hybrid DPO+SFT objective: the DPO component sculpts the fine-tuned policy's ($\pi_{\theta}$) decision boundary to favor low-cost actions, while the SFT anchor preserves high-reward behaviors learned by the expert policy $\pi_{ref}$.

Overall, \textsc{PREFINE} achieves the best balance between reward and safety across all tasks, remaining consistently below the cost threshold. These results highlight \textsc{PREFINE}’s practical deployability, as it yields high-performing yet reliably safe policies across diverse environments. We provide the training curves in Appendix.
\begin{figure}[!htbp]
        \centering
        \includegraphics[width=0.7\linewidth, height = 4.5cm]{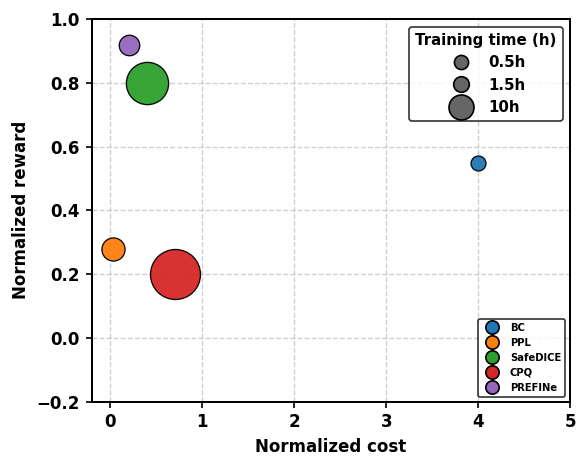}
    \caption{Wall-clock running time (proportional to marker size) comparison of \textsc{PREFINE} with baselines. \textsc{PREFINE} is more scalable.}
    \label{fig:runtime}
\end{figure}
\subsection{Efficiency (Q2)}
Efficiency is critical for deploying safety alignment at scale. Figure \ref{fig:runtime} shows the comparison of \textsc{PREFINE} wall-clock running time with the baselines for the task Walker2dVelocity. The running time is proportional to the marker size. \textsc{\textsc{PREFINE}} achieves a dramatic runtime advantage over competing baselines (e.g., SafeDICE, CPQ). In our experiments \textsc{Prefine} completes end-to-end fine-tuning in about 1.5 hours, roughly an order of magnitude faster (10× speedup) than CPQ and SafeDICE, which take more than 10 hours to finish running. This improvement stems from \textsc{Prefine}'s single-stage preference fine-tuning (one backward pass per minibatch) versus the nested optimization or distribution-matching updates required by SafeDICE/CPQ. To ensure an apples-to-apples comparison all methods were executed on identical hardware (NVIDIA Tesla V100, 20\,GB), used the same dataset splits and random seeds, and shared implementation-level settings (matched network architectures, optimizer, batch size and number of training iterations). We measured wall-clock training time excluding evaluation, used authors' reference code when available or reimplemented baselines using PyTorch with hyperparameters taken from the original papers; these controls ensure the observed timing gap reflects algorithmic cost rather than implementation-based differences.
\small
\begin{figure*}[ht]
  \centering
  \begin{subfigure}[b]{0.48\textwidth}
    \centering
    \includegraphics[width=0.7\linewidth, height = 6cm]{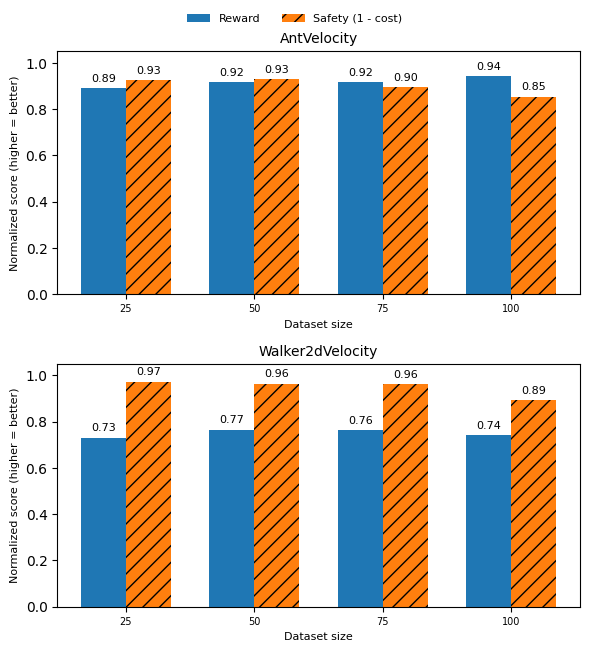}
    \label{fig:sub1}
  \end{subfigure}
  \hfill
  \begin{subfigure}[b]{0.48\textwidth}
    \centering
    \includegraphics[width=0.7\linewidth, height = 6cm]{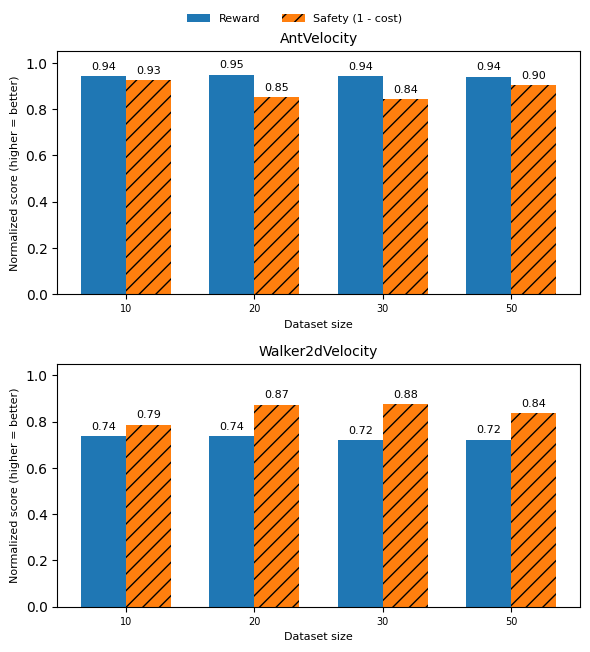}
    \label{fig:sub2}
  \end{subfigure}
  \caption{Robustness of \textsc{PREFINE} to dataset size. \textsc{PREFINE} maintains consistently high normalized rewards and strong safety across varying dataset sizes for $\mathcal{D}_{p}$ (left) and $\mathcal{D}_{np}$ (right), demonstrating stability and data efficiency.}
\label{fig:q3}
\end{figure*}
\begin{figure}[t]
    \centering
    \begin{subfigure}[t]{0.22\textwidth}
        \centering
        \includegraphics[width=\linewidth]{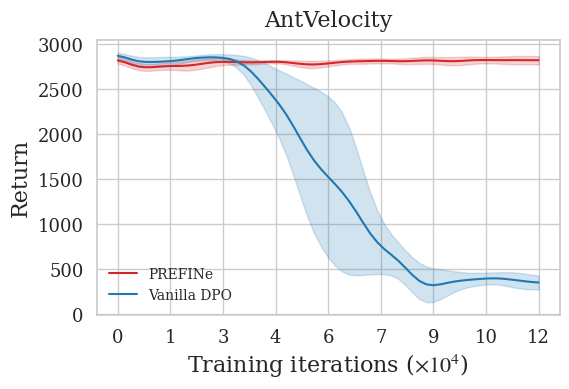}
        \label{fig:ab1_a}
    \end{subfigure}
    \begin{subfigure}[t]{0.22\textwidth}
        \centering
        \includegraphics[width=\linewidth]{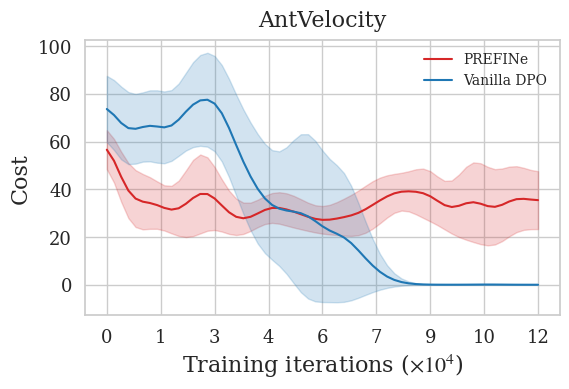}
        \label{fig:ab1_b}
    \end{subfigure}\\
    \begin{subfigure}[t]{0.22\textwidth}
        \centering
        \includegraphics[width=\linewidth]{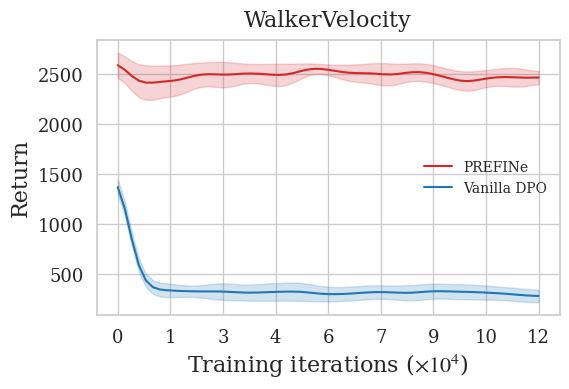}
        \label{fig:ab1_a}
    \end{subfigure}
    \begin{subfigure}[t]{0.22\textwidth}
        \centering
        \includegraphics[width=\linewidth]{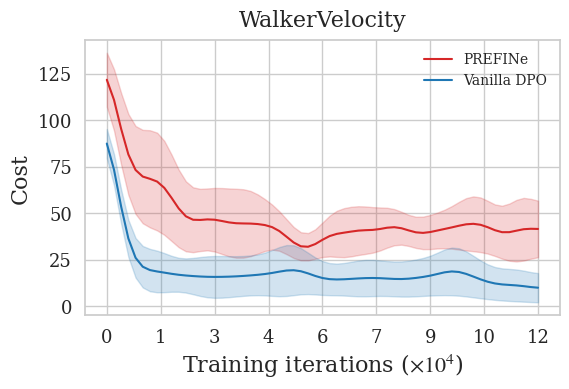}
        \label{fig:ab1_b}
    \end{subfigure}
    \small
    \caption{
        \textbf{Ablation study:} 
        (Left) Safety alignment of reference policy $\pi_{ref}$ using \textsc{PREFINE}(red) vs vanilla DPO loss(blue). \textsc{PREFINE} uses DPO + SFT loss terms. DPO return (left) rapidly falls down due to unlearning, while \textsc{PREFINE} avoids unsafe behavior while retaining high task performance. Vanilla DPO(blue) shows lower cost in comparison to \textsc{PREFINE} (right) but at the same time, struggles to keep the expert ($\pi_{ref}$) performance intact.
    }
    \label{fig:ab1_app}
\end{figure}
\begin{figure}[!htbp]
        \centering
        \includegraphics[width=\linewidth]{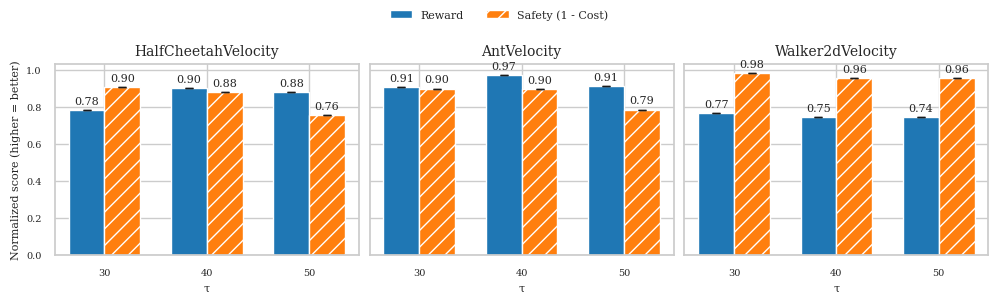} 
    \caption{Robustness of \textsc{PREFINE} to cost thresholds. \textsc{PREFINE} maintains consistently high rewards and strong safety across different values of $\tau$, demonstrating stability.}
    \label{fig:tau}
\end{figure}
\subsection{Robustness \& Sensitivity (Q3)}
A key strength of \textsc{PREFINE} is its stability under different cost thresholds and dataset configurations. In Figure \ref{fig:tau}, we examine the effect of varying the cost threshold $\tau$ across multiple tasks. \textsc{PREFINE} maintains consistently high normalized rewards while safety (measured as 1 - cost) remains robust across $\tau=30,40,50$. The cost slightly increases (and safety decreases) as the cost threshold becomes relaxed which is expected. This indicates that \textsc{PREFINE} does not overfit to a particular safety threshold and reliably balances reward and safety across a wide range of cost threshold values.

We also study \textsc{PREFINE}’s robustness with respect to dataset size, as shown in Figure \ref{fig:q3}. On both AntVelocity and Walker2dVelocity, \textsc{PREFINE} achieves stable performance even with relatively small offline datasets. Both normalized rewards and safety stay consistently high. This demonstrates that \textsc{PREFINE} is data-efficient, able to learn safe and rewarding policies even when training data is limited. This quality is especially useful for real-world scenarios where collecting a large number of unsafe demonstrations for training task agents can be challenging.

Next, we sweep the SFT weight \(\lambda\) across \(\{0.1, 1.0, 1.6, 2.0\}\) and DPO hyperparameter $\beta$ across \(\{0.05, 0.2, 0.6, 0.95\}\) (see Figure \ref{fig:ab5_app} in Appendix). We find that \textsc{PREFINE} performs well for \(\lambda\in{1.0,1.6}\) and \(\beta\in{0.05,0.2}\) on the tasks we tested. In summary, higher value of $\lambda$ and a lower value of $\beta$ work well. Note that $\beta$ controls the influence of the KL-divergence term in vanilla DPO loss thereby deciding the closeness of the fine-tuned policy to the reference policy. For \textsc{PREFINE}, using a large $beta$ with vanilla DPO loss alone was not working well, as evident by the rapid unlearning effect at the beginning of the fine-tuning process (see Section \ref{sec:ablation}).\\
\textbf{Discussion}  
These results illustrate a fundamental shift: rather than laboriously engineering cost functions or performing expensive online interactions, we can leverage naturally occurring preferences to retrofit safety into a differentiable pre-trained policy quickly and robustly. \textsc{PREFINE} thus paves a practical path towards safe deployment in domains where data is not easily available and safety is non-negotiable such as autonomous driving systems (reducing collisions) to healthcare (avoiding dangerous drug interactions). However, policy-sampled counterfactual actions for computing DPO loss can introduce noise when sampled actions are not strictly safer/more unsafe. SFT regularization mitigates but does not eliminate this. Future work should explore ways to mitigate adversarial or erroneous feedback.
\subsection{Ablation Study}
\label{sec:ablation}
We conduct an ablation study for evaluating the impact of the supervised fine-tuning (SFT) loss $\mathcal{L}_{\text{SFT}}$. We compare \textsc{PREFINE} which combines DPO with SFT against vanilla DPO (Figure \ref{fig:ab1_app}). \textbf{Without the SFT term, the policy reduces constraint violations but suffers a sharp drop in task performance, becoming overly conservative. Return collapses below 88\% within the first 10K steps (see Fig.~\ref{fig:ab1_app}).} In contrast, including SFT enables the policy to balance safety and reward, maintaining high returns while enforcing constraints. This highlights SFT as essential for preventing catastrophic unlearning and achieving a robust safety–performance trade-off. Additional ablation results are provided in the Appendix.
\section{Conclusion}
We present \textsc{PREFINE}, a 
fully offline framework for post-hoc safety alignment of pre-trained differentiable policies. By creating counterfactual trajectories from offline data in continuous domains, \textsc{PREFINE} casts safety as a preference learning problem and makes it amenable to Direct Preference Optimization (DPO) and supervised fine-tuning (SFT).
\textsc{PREFINE} does not perform nested optimization and explicit cost estimation. Our experiments diverse on continuous control and navigation benchmarks show that it substantially reduces constraint violations by over 60\% while maintaining expert-level task performance. It runs an order of magnitude faster than the baseline methods, completing end-to-end training in under 1.5 hours. These results underscore \textsc{PREFINE}’s potential as a scalable recipe for retrofitting safety into existing policies and paves the way for fast deployment.
\bibliographystyle{plain}
\bibliography{ref}

\clearpage
\newpage

\appendix
\section{Appendix}
\begin{table*}[h]
\centering
\caption{Preference dataset specifications for all domains used in our experimental results on DSRL Mujoco and Safety Gym tasks.}
\begin{tabular}{lccccccc}
\toprule
\textbf{Task specification} & \textbf{Ant} & \textbf{HalfCheetah} & \textbf{Hopper} & \textbf{Walker2d} & \textbf{Swimmer} & \textbf{PointGoal1} & \textbf{CarGoal1} \\
\midrule
\# of preferred demonstrations ($|\mathcal{D}_p|$)              & 100 & 100 & 100 & 100 & 100 & 100 & 100 \\
\# of non-preferred demonstrations ($|\mathcal{D}_{np}|$)       & 20 & 20 & 20 & 20 & 20 & 20 & 20 \\
Mean cost of preferred demonstrations                           & 26.15 & 23.12 & 26.12 & 10.15 & 24.62 & 9.12 & 9.04 \\
Mean cost of non-preferred demonstrations                       & 225.10 & 221.65 & 225.35 & 273.35 & 174.55 & 88.60 & 108.25 \\
Mean return of preferred demonstrations                         & 2485.10 & 2415.50 & 1616.55 & 2661.96 & 132.4 & 20.30 & 27.03 \\
Mean return of non-preferred demonstrations                     & 2667.70 & 2418.10 & 1157.23 & 2648.11 & 102.6 & 16.47 & 20.42 \\
\bottomrule
\end{tabular}
\label{tab:combined_prefdata}
\end{table*}

\begin{table*}[h]
\centering
\caption{Preference Dataset specification of each domain used in our experimental results on DSRL Bullet Gym tasks.}
\begin{tabular}{lccccc}
\toprule
\textbf{Task specification} & \textbf{BallRun} & \textbf{CarRun} & \textbf{BallCircle} & \textbf{CarCircle} & \textbf{DroneRun} \\
\midrule
Safety Constraint ($\kappa$) & 80 & 40 & 80 & 100 & 140 \\
\# of preferred demonstrations ($|\mathcal{D}_p|$)              & 100 & 100 & 100 & 100 & 100 \\
\# of non-preferred demonstrations ($|\mathcal{D}_{np}|$)       & 20 & 20 & 20 & 20 & 20 \\
Mean cost of preferred demonstrations                           & 6.6 & 34.84 & 12.54 & 15.77 & 11.6 \\
Mean cost of non-preferred demonstrations                       & 74.8 & 34.84 & 79.6 & 92.4 & 134.22 \\
Mean return of preferred demonstrations                         & 275.8 & 530.69 & 740.43 & 361.47 & 387.23 \\
Mean return of non-preferred demonstrations                     & 1021.72 & 505.06 & 458.36 & 259.75 & 481.42 \\
\bottomrule
\end{tabular}
\label{tab:bg}
\end{table*}

\subsection{DSRL Task Description}
We evaluate our approach on the DSRL benchmark ~\cite{liu2024dsrl}, a widely adopted suite for studying offline safe reinforcement learning. DSRL offers a comprehensive set of 38 datasets spanning multiple safety-critical environments and difficulty levels. We choose 12 tasks out of SafetyGymnasium ~\cite{ray2019safetygym, ji2023safetygymnasium} and BulletSafetyGym ~\cite{gronauer2022bulletsafetygym} because of compute limitations. Each environment in DSRL is specifically designed to assess the trade-off between task performance and safety in offline learning. We adopt this benchmark because it provides diverse safety constraint types including collision hazards, velocity limits, and dynamic obstacle avoidance allowing for a systematic and fair comparison across methods under heterogeneous safety formulations.

\textbf{Safety Gym}
Built on the Mujoco physics engine, SafetyGymnasium provides a variety of safety-aware control tasks. It includes two main agent types, \emph{Car} and \emph{Point}, each associated with four task categories: \emph{Button}, \emph{Circle}, \emph{Goal}, and \emph{Push}. These tasks are further divided into difficulty levels (\emph{1} and \emph{2}) and follow the naming pattern \texttt{{Agent}{Task}{Difficulty}}. Agents must navigate toward goals while avoiding hazards. The benchmark also contains five velocity-constrained Mujoco agents: \emph{Ant}, \emph{HalfCheetah}, \emph{Hopper}, \emph{Walker2d}, and \emph{Swimmer} for studying constraint-driven locomotion. We use all 5 velocity-constrained Mujoco tasks because they run faster along with PointGoal1 and CarGoal1 tasks to maintain constraint diversity.

\textbf{Bullet Gym}
Implemented in the PyBullet simulator, BulletSafetyGym follows similar safety principles but features shorter horizons and a broader range of agents. It comprises four agent types: \emph{Ball}, \emph{Car}, \emph{Drone} each performing two task types: \emph{Circle} and \emph{Run}, labeled as \texttt{{Agent}{Task}}. We choose BallRun, BallCircle, CarRun, CarCircle and DroneRun out of these to cover all agent types.

\subsection{Dataset Characteristics}
We conduct experiments on Safety Gym and Bullet Gym tasks from DSRL suite with safety constraints. Table ~\ref{tab:combined_prefdata} shows the Safety Gym tasks, safety constraint settings and the information of dataset for each domain that we used in our experiments. Likewise, Table ~\ref{tab:bg} shows the dataset characteristics for Bullet Gym tasks.

\subsection{Evaluation Metrics}
To evaluate performance, we follow the protocol established in the \textsc{DSRL} benchmark~\cite{liu2024dsrl}, which reports both \emph{normalized reward} and \emph{normalized cost}. 
The normalized reward is computed based on the cumulative returns achieved by the policy, while the normalized cost is defined as:
\begin{equation}
    C_{\text{normalized}} = \frac{C_{\pi} + \varepsilon}{\kappa + \varepsilon},
\end{equation}
where $C_{\pi}$ denotes the cumulative cost under policy $\pi$, $\kappa$ is the predefined cost threshold, and $\varepsilon$ is a small positive constant to ensure numerical stability when $\kappa = 0$. 
Following the \textsc{DSRL} convention, a task is regarded as \emph{safe} when $C_{\text{normalized}} \leq 1$.

\subsection{Training Details and Hyperparameters}
We adopt a two-step training procedure for PREFINE. First, we pretrain
the reference policy $\pi_{ref}$ using behavior cloning (BC) on high reward trajectories. Next, we refine the policy by applying PREFINE on the newly created preferred and non-preferred datasets. The hyperparameters used
in the experiments are summarized below:

\textbf{Hyperparameters}:
\begin{itemize}
    \item $\gamma$ (discount factor): 0.99
    \item Optimizer: Adam
    \item Learning rate: $3e-4$
    \item Network size: [256,256]
    \item Batch size: 256
    \item Training iterations: 500,000
    \item Hardware: NVIDIA Tesla V100 GPU
    \item Fine-tuning duration: $\sim$1.5 hours per task
\end{itemize}

\textbf{Other Hyperparameters}:
\begin{itemize}
    \item SFT weight $\lambda \in \{0.1, 1.0, 1.6, 2.0\}$
    \item DPO temperature $\beta \in \{0.05, 0.2, 0.6, 0.95\}$
    \item Final values used: $\lambda = 1.6$, $\beta = 0.05$ for Safety Gym; $\lambda = 1.0$, $\beta = 0.2$ for Bullet Gym.
\end{itemize}

\textbf{Baseline Hyperparameters:} We choose the hyperparameters for PPL, SafeDICE and CQL from their respective papers.

\section{Additional Results}
\begin{figure}[!htbp]
        \centering
        \includegraphics[width=\linewidth, height = 6cm]{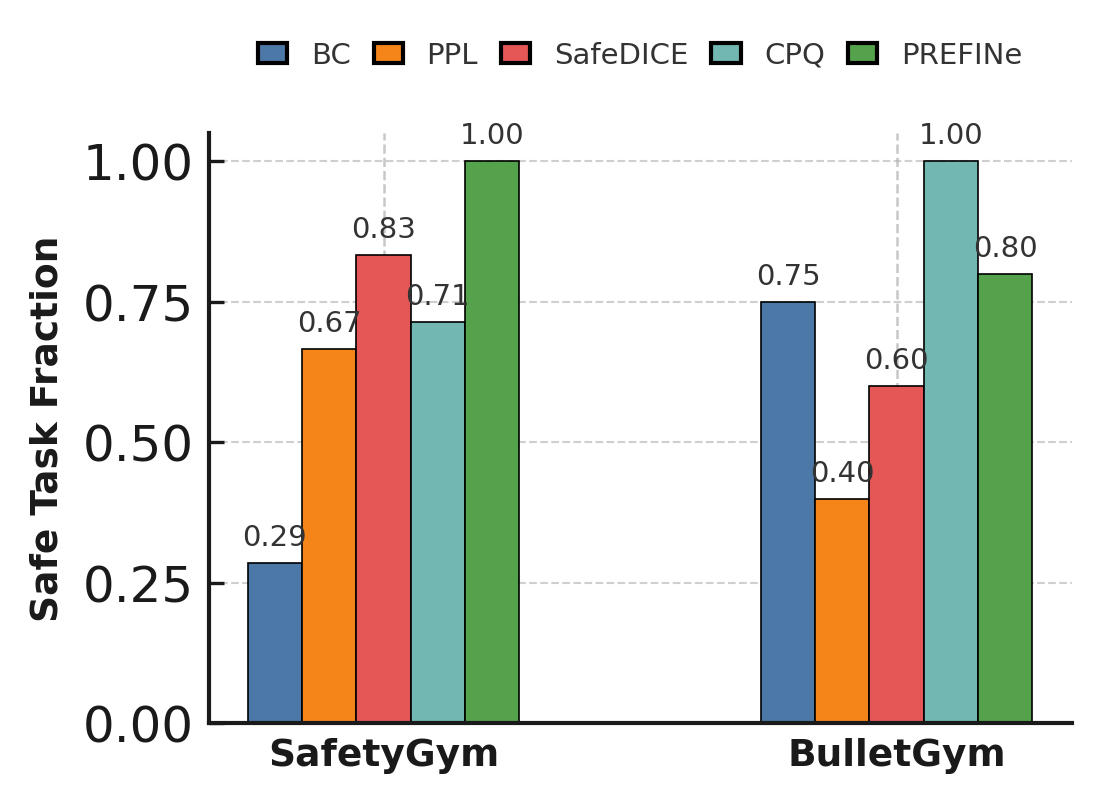}
    \caption{Fraction of tasks solved for safety.}
    \label{fig:safe_percent}
\end{figure}

\subsection{Fraction of Tasks Solved for Safety}
We see in Figure \ref{fig:safe_percent} that PREFINE solves the maximum percentage of tasks safely in both Safety Gym and Bullet Gym task suites. In Bullet Gym, it is second only to CPQ and that is because CPQ trades off rewards for safety.

\subsection{Ablation Study: \textsc{PREFINE} vs. Vanilla DPO}

\begin{figure*}[h]
    \centering
    \begin{subfigure}[t]{0.38\textwidth}
        \centering
        \includegraphics[width=\linewidth]{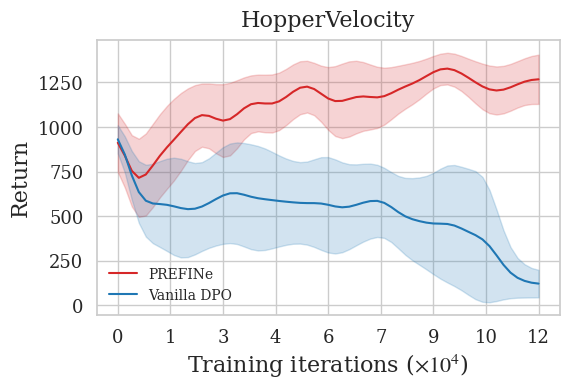}
        \label{fig:ab1_a}
    \end{subfigure}
    \begin{subfigure}[t]{0.38\textwidth}
        \centering
        \includegraphics[width=\linewidth]{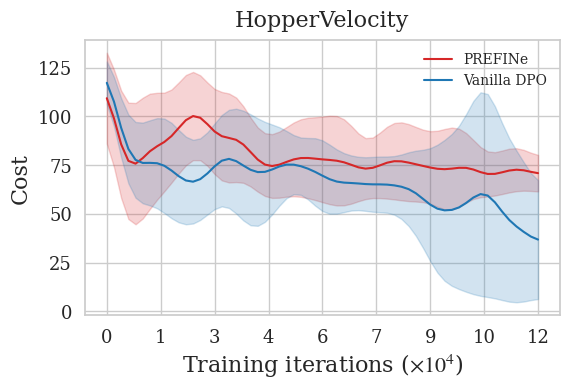}
        \label{fig:ab1_b}
    \end{subfigure}
    \caption{
        \textbf{Ablation study 1:} 
        (Left) Safety alignment of reference policy $\pi_{ref}$ using \textsc{PREFINE}(red) vs DPO loss(blue). \textsc{PREFINE} uses DPO + SFT loss terms. DPO return rapidly falls down due to unlearning, while \textsc{PREFINE} avoids unsafe behavior while retaining high task performance.
        (Right) DPO(blue) shows lower cost in comparison to \textsc{PREFINE} but at the same time, struggles to keep the expert ($\pi_{ref}$) performance intact.
    }
    \label{fig:ab1_hop}
\end{figure*}

Vanilla DPO only ensures relative ranking, so it can deviate heavily from expert behavior if it finds safer modes. This leads to catastrophic unlearning—the policy becomes overly conservative or unstable. It is evident by the returns crashing for Vanilla DPO. The SFT term anchors the policy to expert trajectories, acting like a soft KL term and ensuring the task performance doesn't get sacrificed while improving safety. We show the results for Hopper in Figure \ref{fig:ab1_hop}

\subsection{Effect of \texorpdfstring{$\lambda$}{lambda} and \texorpdfstring{$\beta$}{beta} on Safety Performance Trade-off in Walker}
\small
\begin{figure*}[t]
    \centering
    \begin{subfigure}[t]{\textwidth}
        \centering
        \includegraphics[width=0.9\linewidth]{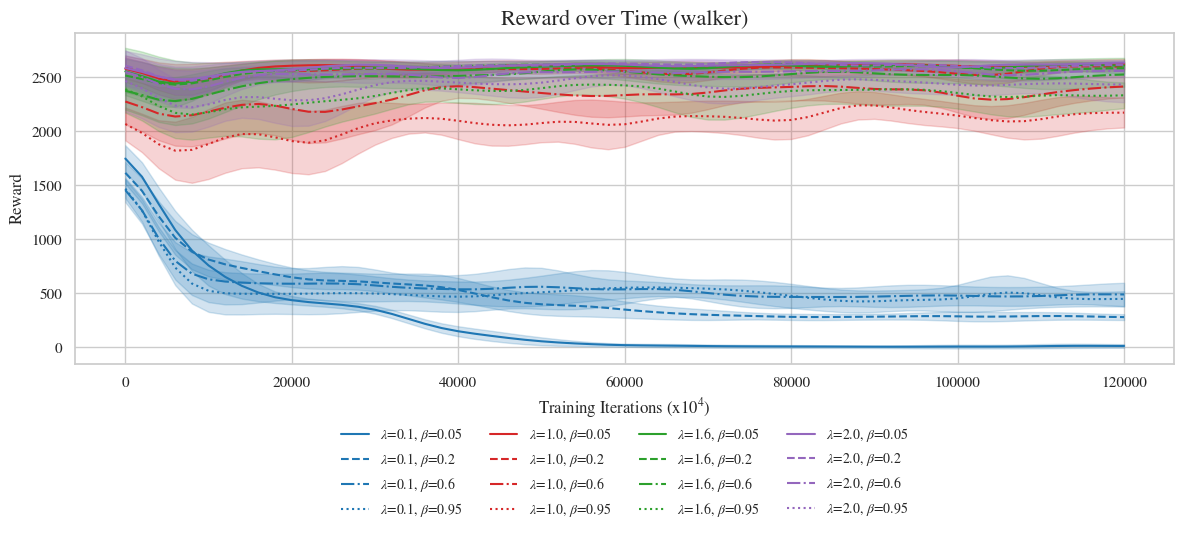}
        \label{fig:ab1_a}
    \end{subfigure}
    \hfill
    \begin{subfigure}[t]{\textwidth}
        \centering
        \includegraphics[width=0.9\linewidth]{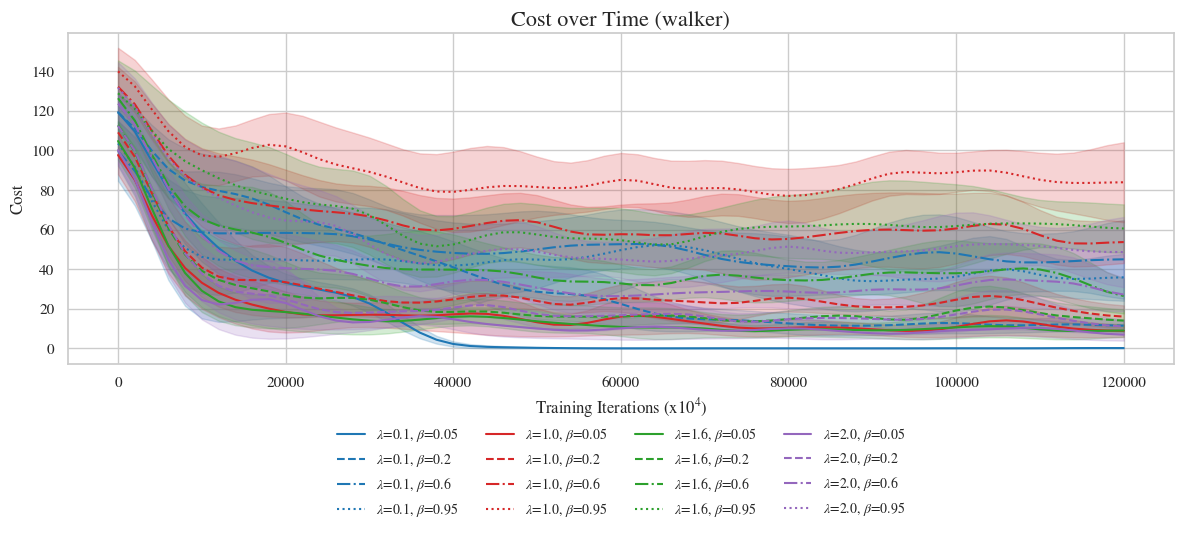}
        \label{fig:ab1_b}
    \end{subfigure}
    \hfill
    \begin{subfigure}[t]{\textwidth}
        \centering
        \includegraphics[width=0.9\linewidth]{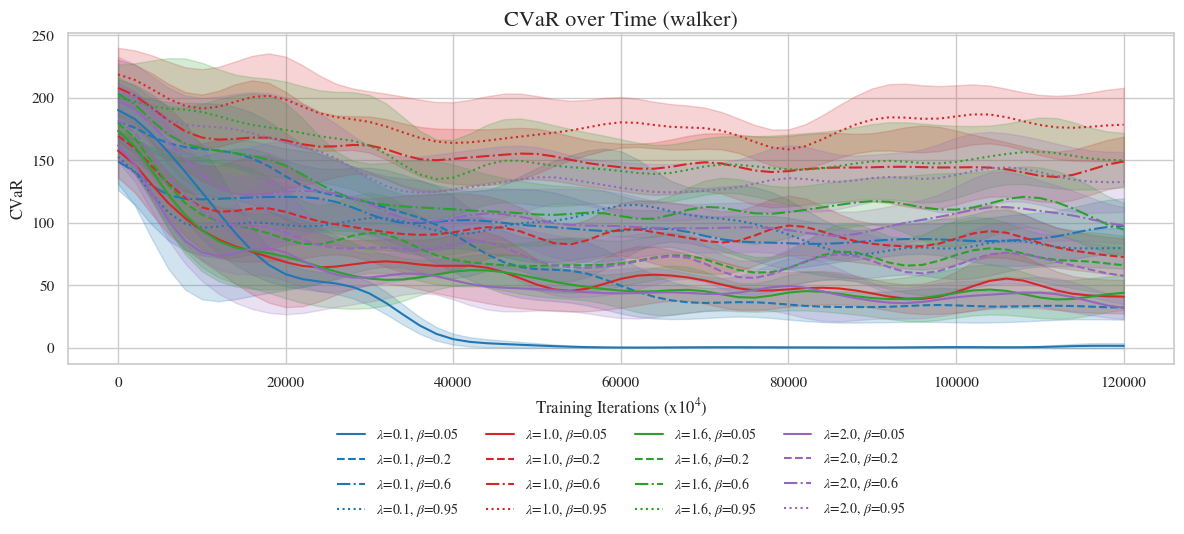}
        \label{fig:ab1_c}
    \end{subfigure}
    \small
    \caption{
        \textbf{Training dynamics for different values of $\lambda$ (color) and $\beta$ (line style) in the Walker environment.} 
        Each subplot shows the evolution of a key metric over training iterations: \textbf{(Top)} cost, \textbf{(Middle)} CVaR, and \textbf{(Bottom)} reward.
        Lower $\lambda$ emphasizes safety, resulting in reduced cost and CVaR but lower reward, especially at higher $\beta$ (e.g., $\lambda = 0.1$, $\beta = 0.95$). 
        Higher $\lambda$ prioritizes reward, often at the expense of safety. Shaded regions denote standard error across three seeds. 
        These plots illustrate the trade-off between safety and performance governed by the $(\lambda, \beta)$ configuration.
    }
    \label{fig:ab5_app}
\end{figure*}

The results in Figure \ref{fig:ab5_app} demonstrate a clear trade-off between safety and performance governed by the choice of $\lambda$ and $\beta$. Lower values of $\lambda$ yield significantly lower cost and CVaR, especially for larger $\beta$, but at the expense of reduced reward. Conversely, higher $\lambda$ improves reward but results in higher safety violations. This highlights the importance of appropriately tuning $(\lambda, \beta)$ to balance safety and performance objectives.

\subsection{Learning Curves}
\begin{figure*}[ht]
  \centering

  \begin{subfigure}[b]{0.45\textwidth}
    \centering
    \includegraphics[width=\linewidth]{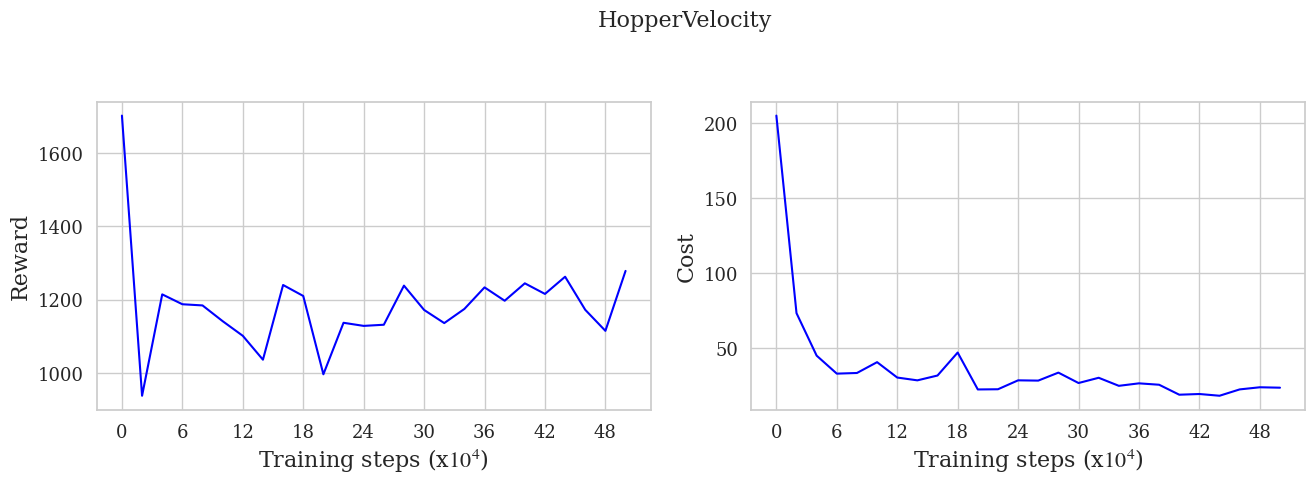}
    \label{fig:sub1}
  \end{subfigure}
  \begin{subfigure}[b]{0.45\textwidth}
    \centering
    \includegraphics[width=\linewidth]{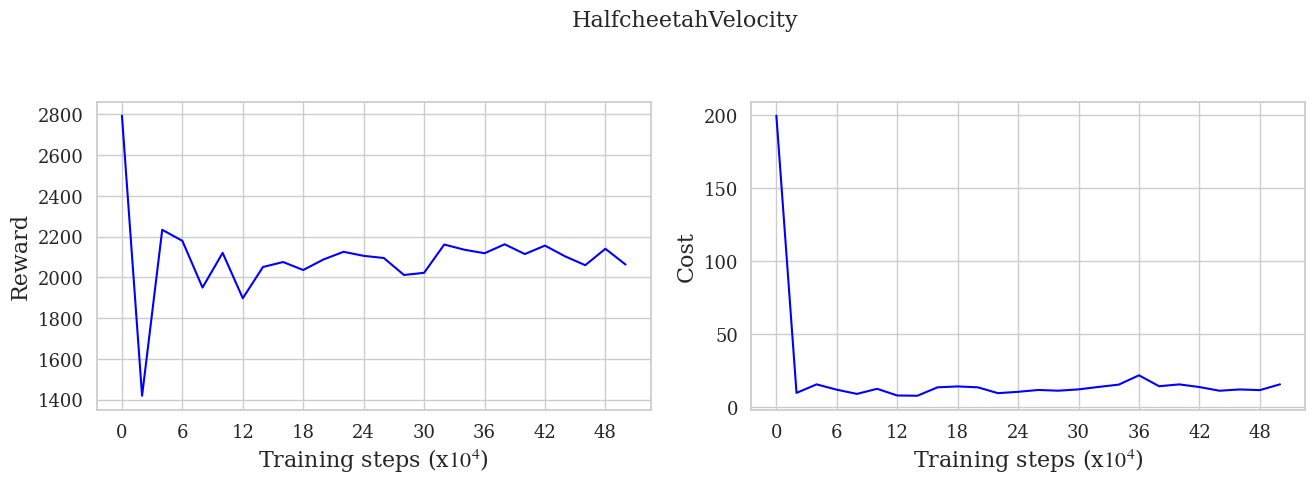}
    \label{fig:sub2}
  \end{subfigure}
  \hfill
  \begin{subfigure}[b]{0.45\textwidth}
    \centering
    \includegraphics[width=\linewidth]{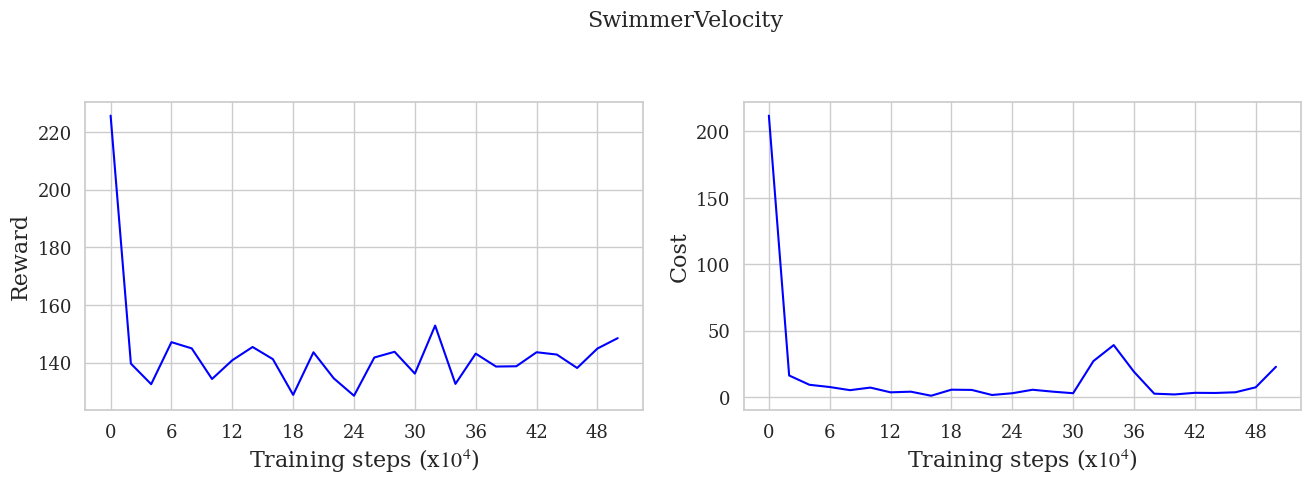}
    \label{fig:sub3}

  \end{subfigure}
  \begin{subfigure}[b]{0.45\textwidth}
    \centering
    \includegraphics[width=\linewidth]{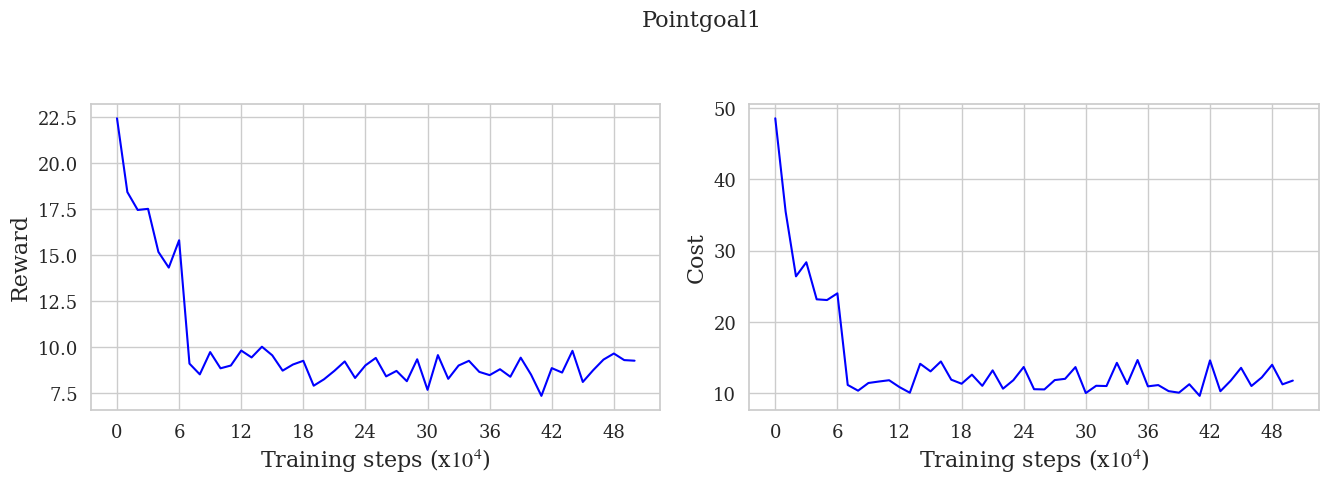}
    \label{fig:sub3}

  \end{subfigure}
  \hfill
  \begin{subfigure}[b]{0.45\textwidth}
    \centering
    \includegraphics[width=\linewidth]{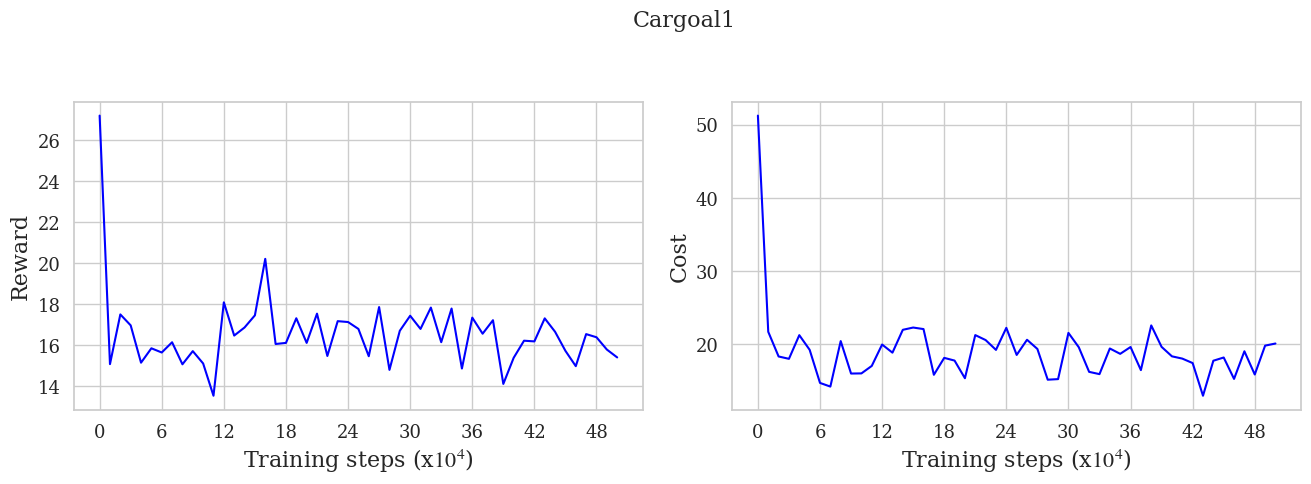}
    \label{fig:sub3}
     \end{subfigure}
\begin{subfigure}[b]{0.45\textwidth}
    \centering
    \includegraphics[width=\linewidth]{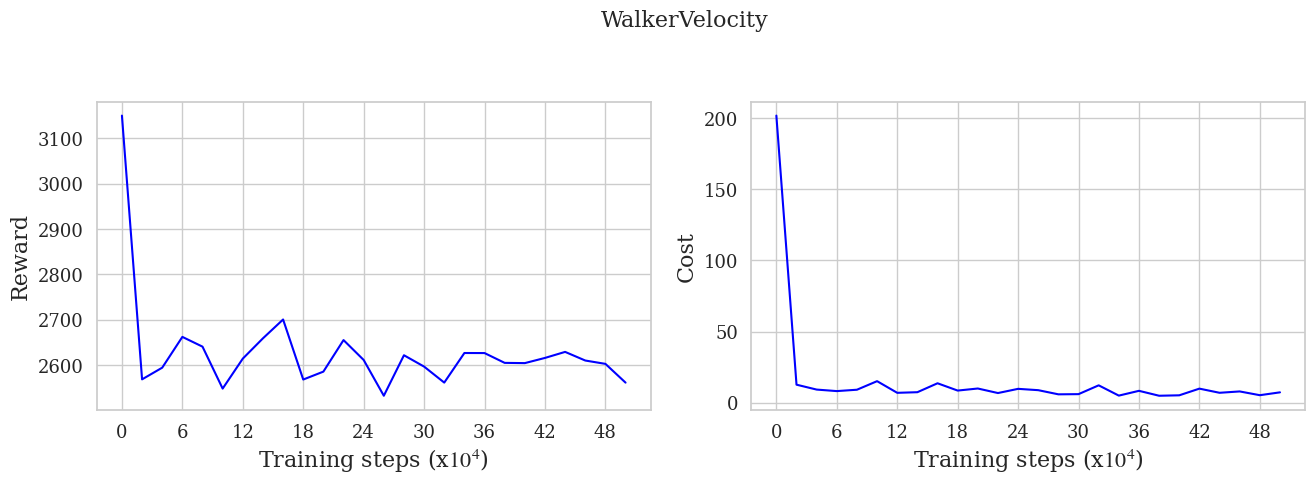}
    \label{fig:sub3}

  \end{subfigure}
  \hfill
  \begin{subfigure}[b]{0.45\textwidth}
    \centering
    \includegraphics[width=\linewidth]{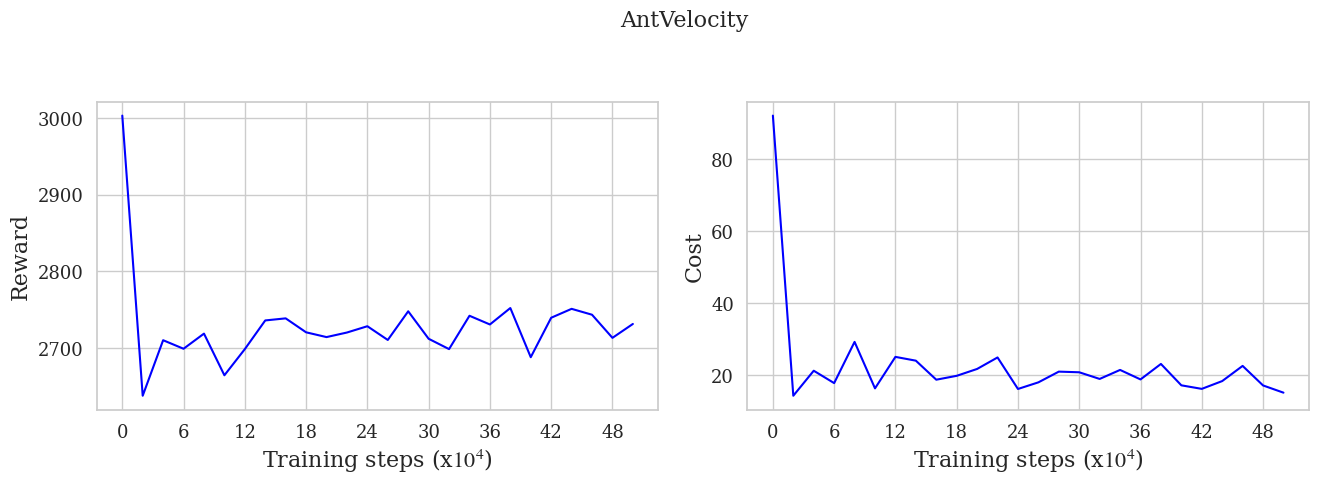}
    \label{fig:sub3}

  \end{subfigure}
 \caption{PREFINE training curves for Safety Gym tasks.}
  \label{fig:sg}
\end{figure*}

\begin{figure*}[ht]
  \centering

  \begin{subfigure}[b]{0.45\textwidth}
    \centering
    \includegraphics[width=\linewidth]{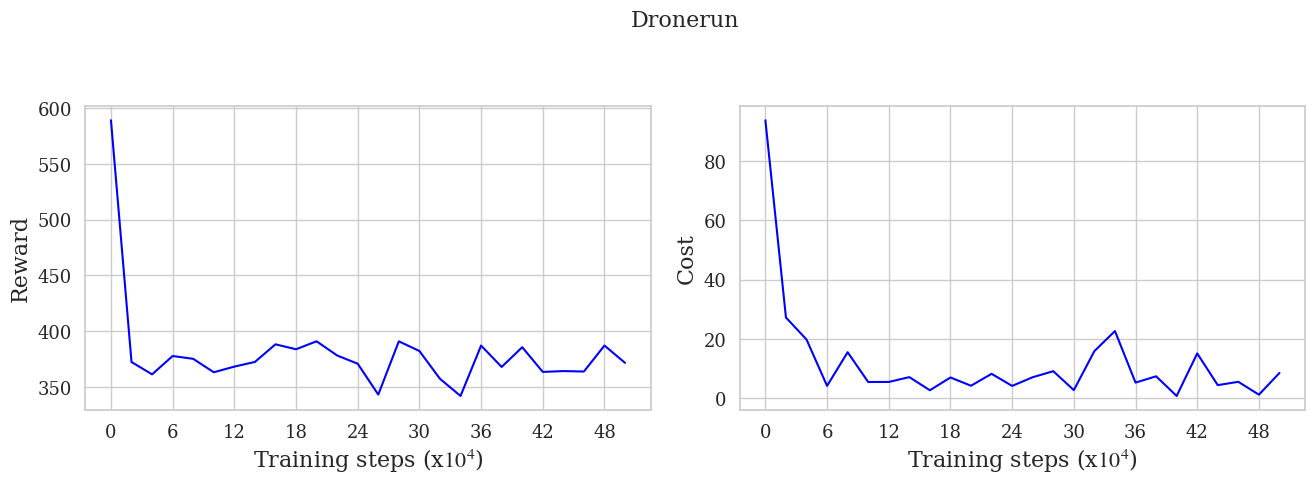}
    \label{fig:sub1}
  \end{subfigure}
  \begin{subfigure}[b]{0.45\textwidth}
    \centering
    \includegraphics[width=\linewidth]{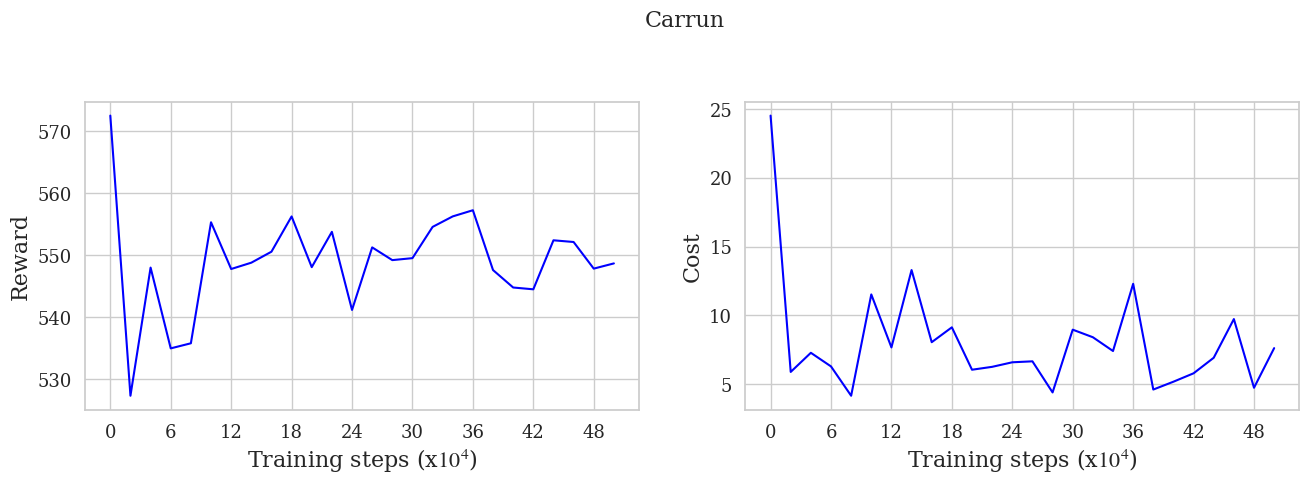}
    \label{fig:sub2}
  \end{subfigure}
  \hfill
  \begin{subfigure}[b]{0.45\textwidth}
    \centering
    \includegraphics[width=\linewidth]{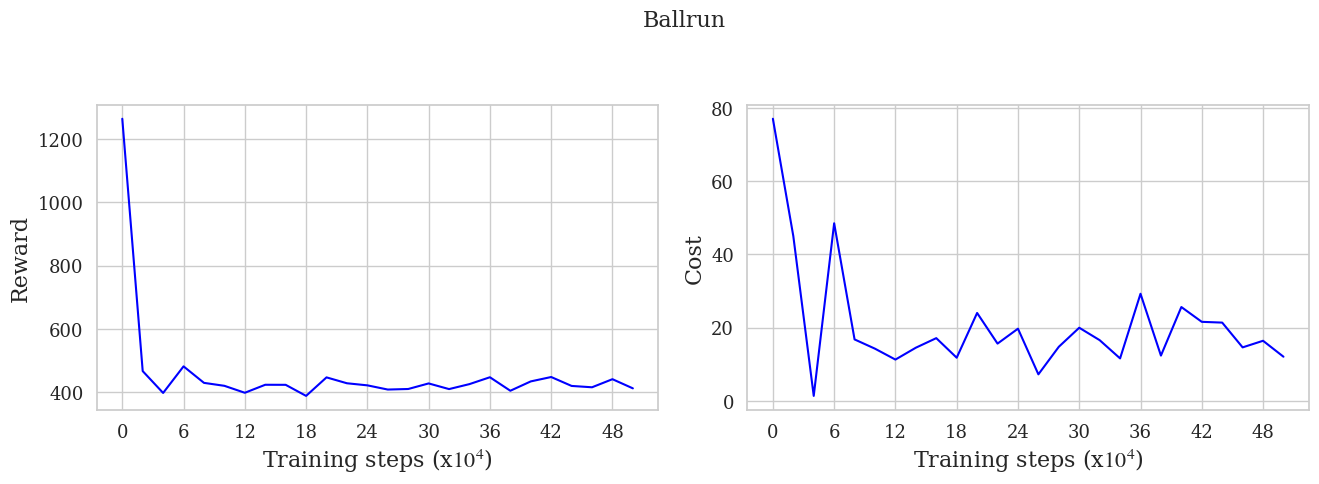}
    \label{fig:sub3}

  \end{subfigure}
  \begin{subfigure}[b]{0.45\textwidth}
    \centering
    \includegraphics[width=\linewidth]{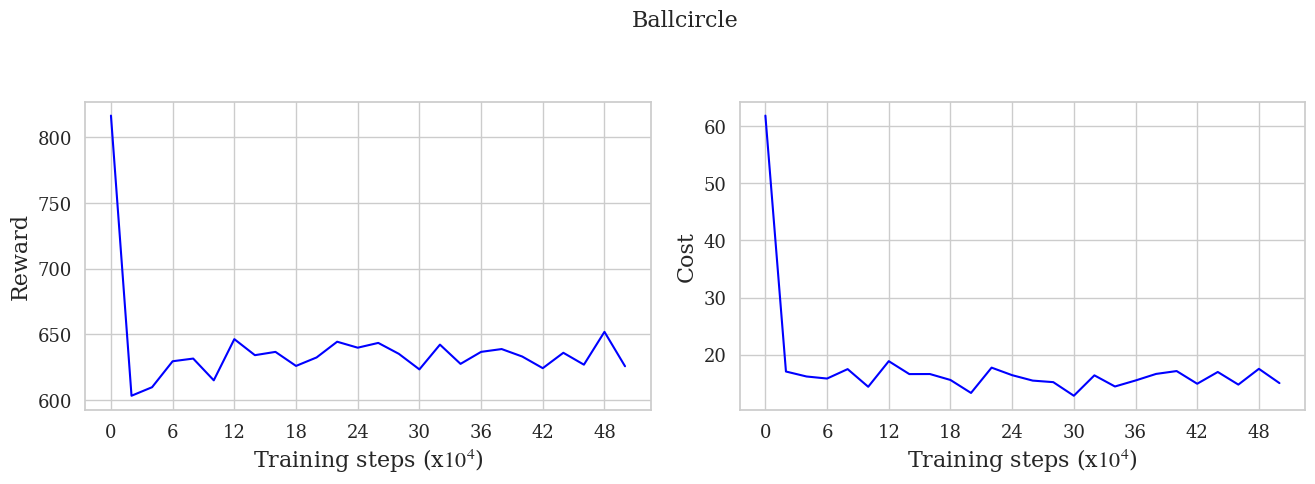}
    \label{fig:sub3}

  \end{subfigure}
  \hfill
  \begin{subfigure}[b]{0.45\textwidth}
    \centering
    \includegraphics[width=\linewidth]{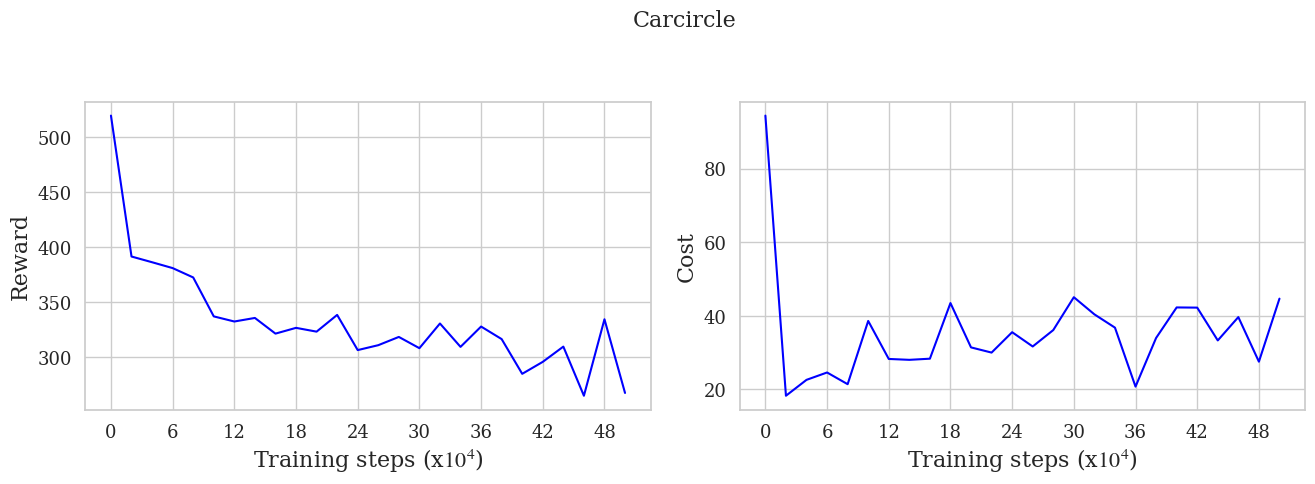}
    \label{fig:sub3}

  \end{subfigure}

 \caption{PREFINE training curves for Bullet Gym tasks.}
  \label{fig:bg}
\end{figure*}

We provide the learning curves for PREFINE (averaged across three cost threshold values: 30, 40 and 50) in Figure . The results are based on absolute rewards and absolute costs. See Figure \ref{fig:sg} for Safety Gym tasks and Figure \ref{fig:bg} for Bullet Gym tasks.

\subsection{Comparison of Policy-based and Dataset-based Sampling Strategies to Generate Counterfactual Actions}For HalfCheetahVelocity and Walker2dVelocity, we observe policy-sampled counterfactuals produce substantially fewer noisy or contradictory preference labels than the mixed sampling strategy (dataset nearest neighbour sampling + policy fallback when nearest neighbour is not found). Figure ~\ref{fig:compare} shows per-quartile counts and percentages of cost-based binary label mismatches for HalfCheetah and Walker. PREFINe has both lower initial mismatch rates and a clearer downward trend during training, whereas the mixed strategy produces markedly higher mismatch rates throughout training (roughly 1.3–1.8× higher). We hypothesize this arises because dataset-sampled counterfactuals introduce stale or distributional-mismatch actions into comparisons: when the dataset contains non-preferred actions that look plausible under current policy features, mixing them into pairwise comparisons leads to contradictory labels and label noise. This label noise corrupts the preference signal and prevents the policy from consistently moving away from unsafe behaviors, explaining the higher empirical cost and CVaR observed for the mixed strategy. In contrast, PREFINe’s policy-sampled counterfactuals produce more coherent comparisons (albeit still noisy in the beginning), enabling more reliable preference gradients and improved safety outcomes.

\subsection{Dataset Overlap}
\begin{figure*}[ht]
  \centering

  \begin{subfigure}[b]{0.3\textwidth}
    \centering
    \includegraphics[width=\linewidth]{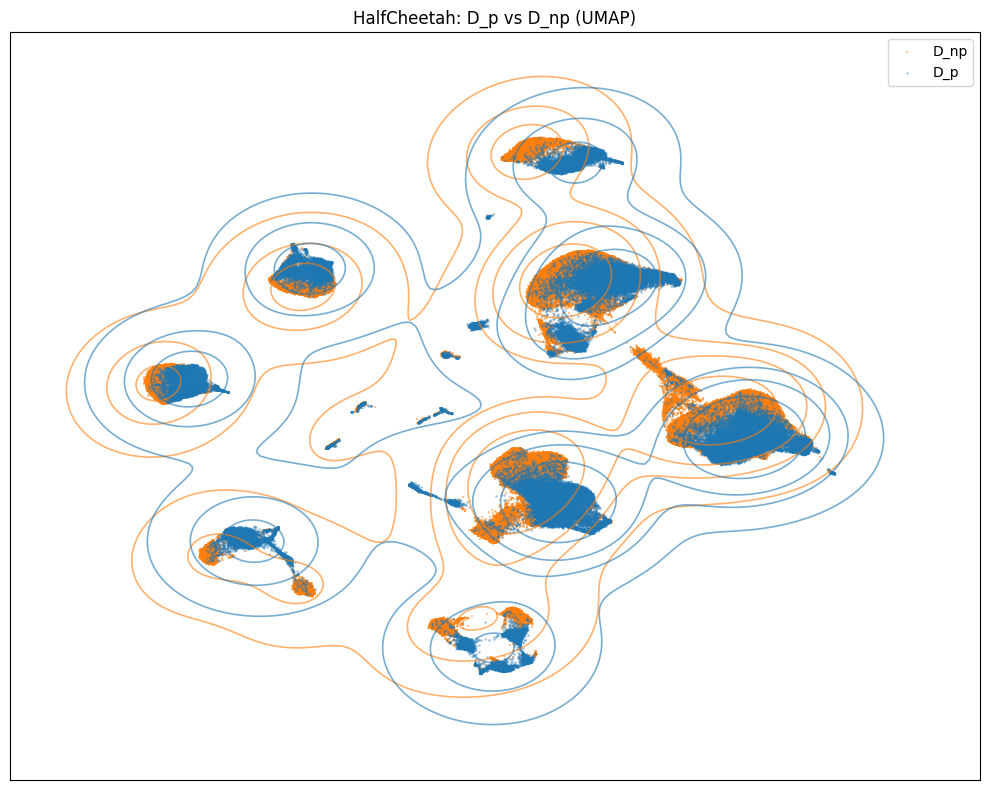}
    \label{fig:sub1}
  \end{subfigure}
  \hfill
  \begin{subfigure}[b]{0.3\textwidth}
    \centering
    \includegraphics[width=\linewidth]{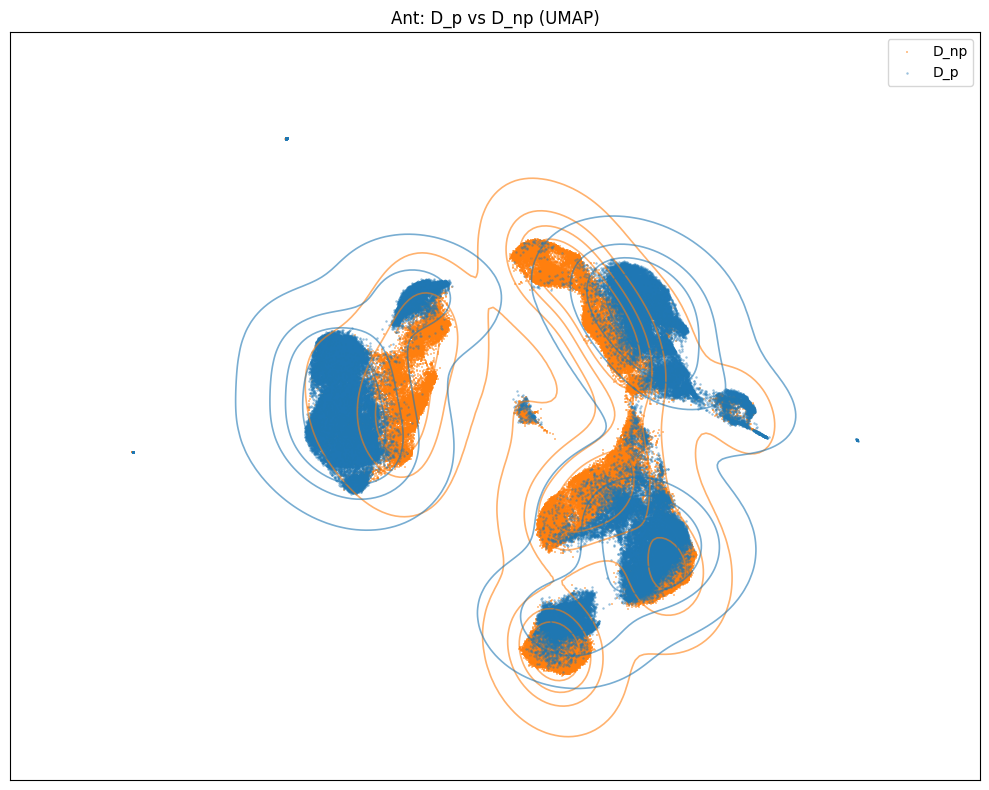}
    \label{fig:sub2}
  \end{subfigure}
  \hfill
  \begin{subfigure}[b]{0.3\textwidth}
    \centering
    \includegraphics[width=\linewidth]{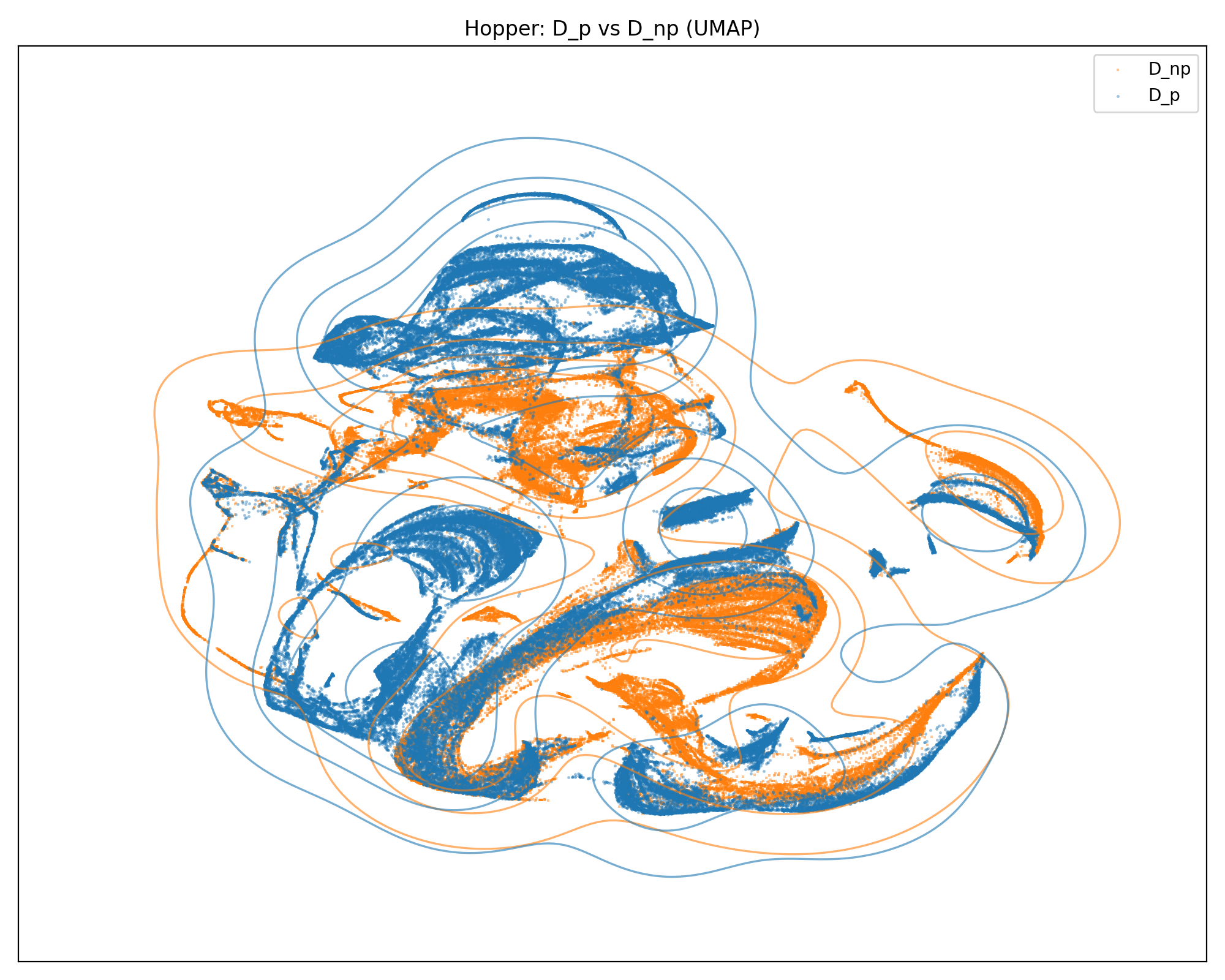}
    \label{fig:sub3}
  \end{subfigure}

  \caption{UMAP embeddings of the training datasets used for various tasks showing significant overlap between Preferred dataset states (blue) and non-preferred dataset states (orange).}
  \label{fig:umap}
\end{figure*}

InFigure ~\ref{fig:umap}, we show the overlap between preferred and dispreferred dataset states. We find that for all the tasks, the Jaccard similarity between the UMAP embeddings of preferred and non-preferred states comes out to be around 70\%. This gives us sufficient confidence to perform counterfactual action sampling from the current policy for \textsc{PREFINE}.

\end{document}